\PassOptionsToPackage{table}{xcolor}
\documentclass[11pt]{article}
\usepackage[preprint]{acl}
\usepackage{times}
\usepackage{latexsym}
\usepackage[T1]{fontenc}
\usepackage[utf8]{inputenc}
\usepackage{microtype}
\usepackage{inconsolata}
\usepackage{graphicx}
\usepackage{amsmath}
\usepackage{mathtools}
\usepackage{booktabs}
\usepackage{multirow}
\usepackage{makecell}
\usepackage{longtable}
\usepackage{amssymb}

\definecolor{bestblue}{RGB}{220,235,247}
\definecolor{refgray}{RGB}{230,230,230}

\newcommand{\best}[1]{%
  \begingroup
  \setlength{\fboxsep}{1.2pt}%
  \colorbox{bestblue}{\strut\textbf{#1}}%
  \endgroup
}

\title{\textsc{SkillBrowser}: Selective Skill Exposure for Scalable Skill-Using Agents}

\author{Jin Ye \\
  Nanjing University \\
  School of Information \\
  Science and Technology \\
  \texttt{yej@smail.nju.edu.cn} \\\And
  Yifan Jiang \\
  Nanjing University \\
  School of Information \\
  Science and Technology \\
  \texttt{231880465@smail.nju.edu.cn} \\\And
  Lan-zhe Guo \\
  Nanjing University \\
  School of Information \\
  Science and Technology \\
  guolz@nju.edu.cn}

\begin{document}
\maketitle
\begin{abstract}
Large language model agents increasingly rely on reusable skills to extend their capabilities in complex tasks. However, scalable skill use requires \emph{Selective Skill Exposure}: providing relevant procedural content when needed while avoiding unnecessary exposure that may disrupt reasoning. Existing paradigms fall short in complementary ways. Progressive disclosure exposes metadata for a growing skill library, introducing context pollution; retrieval limits catalog exposure, but static retrieval from the initial task query may be misaligned with capability needs that emerge during reasoning and may still inject unnecessary skills. We propose \textsc{SkillBrowser}, a Meta-Skill-managed framework that enables an agent to identify current capability needs, formulate targeted queries, and selectively load skills on demand. Across six SRA-Bench subsets and five LLMs, \textsc{SkillBrowser} improves aggregate accuracy over static retrieval and LLM-based selection, including multi-step tool use, and mitigates accuracy degradation on already-solvable instances. Benefits remain model- and task-dependent, with under-invocation of useful skills and additional latency as key limitations. These results demonstrate the viability of Meta-Skill-managed access for scalable skill repositories.
\end{abstract}

\section{Introduction}

\begin{figure*}[t]
    \centering
    \includegraphics[width=\textwidth]{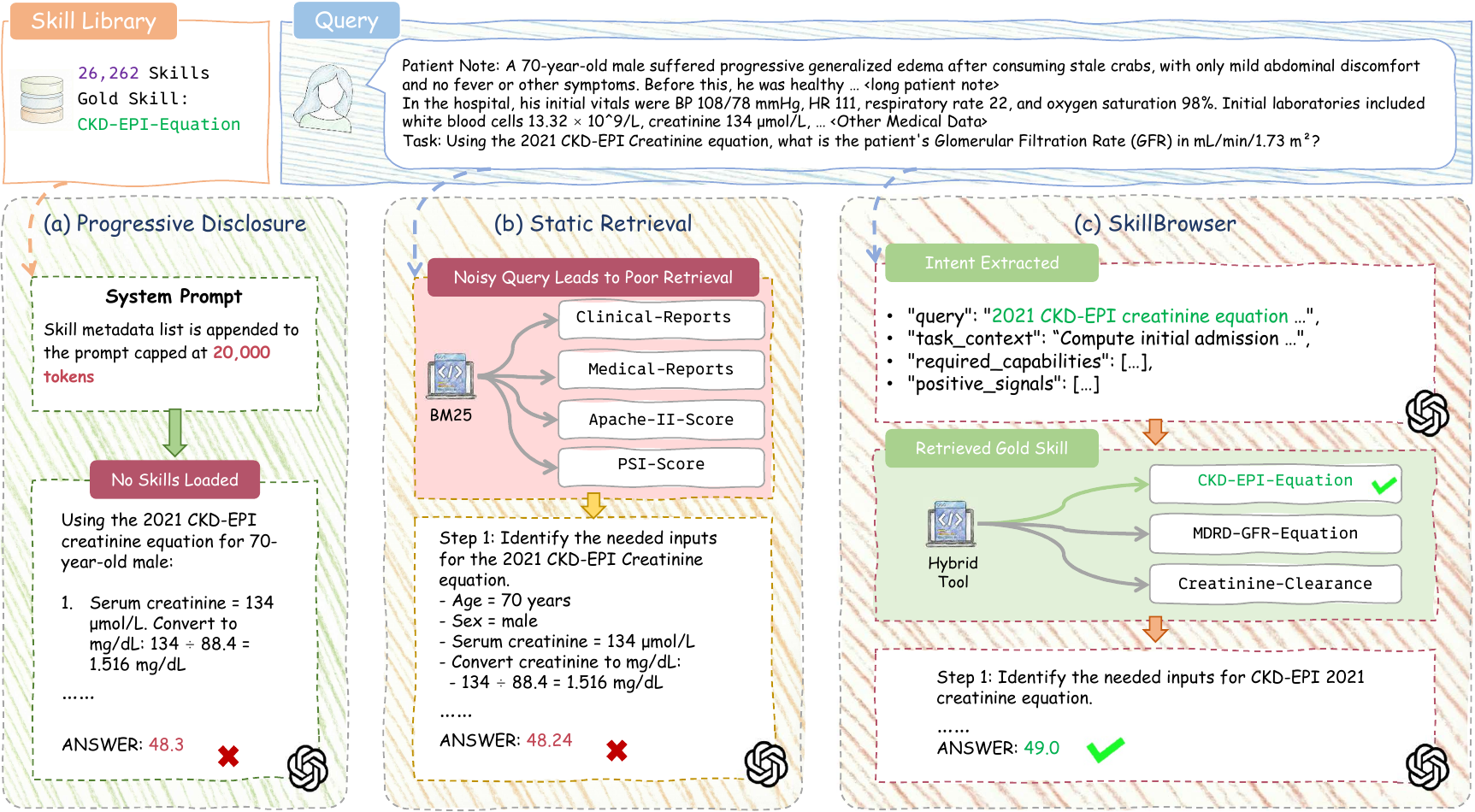}
    \caption{Motivating example of Skill access over a large repository. (a) Progressive disclosure exposes excessive Skill metadata to the agent and distracts it from loading the required Skill, while (b) static retrieval is misled by a noisy user query. (c) \textsc{SkillBrowser} lets the agent formulate a capability-oriented query and retrieve the required Skill on demand, enabling a correct solution.}
    \label{fig:motivating_example}
\end{figure*}

Large language models have rapidly evolved from passive text generators into agentic systems capable of planning, tool invocation, environmental interaction, and iterative problem solving. As LLM-based agents are applied to increasingly open-ended and long-horizon tasks, they must draw on external reusable capabilities beyond their static parametric knowledge. Such modularity enables agents to reuse procedural knowledge and interact with task-specific resources in a structured manner.

Skills extend LLM agents with reusable task-specific capabilities: each Skill contains procedural instructions and may include auxiliary resources such as scripts or references~\citep{Agent2026xu}. Unlike callable tools, loaded Skills contribute instructions directly to the agent's working context and can therefore influence its subsequent reasoning~\citep{li2026skillsbenchbenchmarkingagentskills, han2026sweskillsbenchagentskillsactually}. This property raises a central scalability problem, which we term \emph{Selective Skill Exposure} (SSE): a scalable Skill system should expose the procedural support needed for solving a task while avoiding unnecessary Skill information that may disrupt the agent's reasoning.

Existing paradigms fall short of SSE in complementary ways. Recent Skill-compatible agents commonly adopt progressive disclosure, exposing lightweight metadata, typically Skill names and descriptions, before loading detailed instructions and resources on demand~\citep{Agent2026xu, sokagenticskills2026jiang}. Although this avoids preloading every full Skill, its globally exposed metadata still grows with the repository, consuming context budget, introducing semantic noise, and preventing the agent from accessing the required Skill~\citep{SkillRouter2026zheng}. As illustrated in Figure~\ref{fig:motivating_example} (a), metadata exposure alone may exhaust the available context before the Gold Skill is loaded.

Retrieval provides a natural alternative by selecting a small subset of Skills from an external repository~\citep{SRA2026su,hongcheol_2026,lumer2025scalemcpdynamicautosynchronizingmodel}. However, static Skill retrieval addresses exposure size rather than SSE itself. First, retrieving from the initial user request may be misaligned with the procedural capability required during solving: as shown in Figure~\ref{fig:motivating_example} (b), that capability may be obscured by a noisy request or emerge only after reasoning begins. Second, once retrieved, Skill contents are injected irrespective of whether external procedural assistance is necessary. Because Skills alter the reasoning context, such blind injection may interfere with tasks that the agent could otherwise solve correctly.

To address this problem, we propose \textsc{SkillBrowser}, a Meta-Skill-managed framework for selective Skill exposure over large Skill repositories. Rather than globally exposing Skill metadata or unconditionally injecting statically retrieved Skills, \textsc{SkillBrowser} equips the agent with a reusable Meta-Skill that governs access to the repository during reasoning. The agent can identify its current capability need, generate a targeted query, retrieve relevant Skills through \textsc{SkillGrep}, and load selected Skill contents only on demand, as illustrated in Figure~\ref{fig:motivating_example} (c). This design changes repository access from passive exposure to agent-controlled acquisition: repository scale is decoupled from the context visible to the agent, while unnecessary procedural content need not enter an already successful reasoning process.

Our contributions are as follows:
\begin{itemize}
    \item We formulate \emph{Selective Skill Exposure} as a central problem in scaling Agent Skills, and identify two complementary failures of existing paradigms: metadata-level context pollution under progressive disclosure and capability misalignment or unnecessary Skill exposure under static retrieval.
    
    \item We propose \textsc{SkillBrowser}, a Meta-Skill-managed framework that allows agents to decide whether external Skills are needed, formulate structured capability-oriented queries and load retrieved Skill contents on demand during problem solving.
    
    \item We develop \textsc{SkillGrep}, a capability-oriented Skill search system that expands Skill metadata into multiple retrieval views and applies hybrid retrieval to align agent-formulated needs with relevant Skills.
    
    \item We conduct experiments on SRA-Bench~\citep{SRA2026su}, showing improved end-task performance over static retrieval and substantially reduced degradation from unnecessary Skill exposure on already-solvable instances.
\end{itemize}

\section{Related Work}

\subsection{Skill-Based Agents}

Early tool-augmented agents extend LLMs with external callable capabilities: ReAct~\citep{ReAct2022yao} establishes the reasoning-and-acting paradigm, while API-Bank, ToolLLM, and Gorilla~\citep{APIBank2023li, ToolLLM2024qin, Gorilla2024patil,patel2026dynamictooldependencyretrieval} study API use and selection over increasingly large tool collections. Recent Agent Skills broaden this abstraction to reusable capability packages containing procedural instructions and optional resources such as scripts, templates, examples, or references~\citep{Agent2026xu}. Unlike callable tools, Skills are loaded as procedural context for subsequent reasoning.

To access Skills efficiently, recent systems commonly adopt progressive disclosure, exposing compact metadata before loading detailed Skill contents on demand~\citep{Agent2026xu,sokagenticskills2026jiang}. Although this avoids preloading full Skill documents, the exposed metadata catalog still grows with the repository and may hinder reliable Skill selection at scale~\citep{SkillRouter2026zheng}. This limitation motivates retrieving a compact candidate set rather than exposing the full Skill catalog; however, as discussed next, retrieval alone does not determine whether the returned procedural contents should enter the agent's reasoning context.

\subsection{Skill Retrieval Methods}

Retrieval provides an alternative to global capability exposure. Prior work investigates dedicated tool retrieval, feedback-enhanced retrieval, and generation-based tool selection over large tool inventories~\citep{Enhancing2024xu,ToolGen2025wang, ToolRet2025shi,fei2025mcpzeroactivetooldiscovery}. ToolReAGt further performs iterative retrieval of callable tools for decomposed complex requests~\citep{ToolReAGt2025braunschweiler}. These methods are related to our work in external capability access, but focus on executable tool interfaces rather than procedural Skill contents injected into the model context.

In the Skill setting, SRA retrieves Skills from a large external corpus and evaluates both retrieval quality and downstream Skill use~\citep{SRA2026su,liu2026graphskillsdependencyawarestructural}, while SkillRouter studies large-scale Skill routing with richer Skill representations~\citep{SkillRouter2026zheng}. We study a complementary problem: static retrieval may fail to match capability needs that emerge during reasoning, and retrieved Skill contents need not always be exposed to the agent. \textsc{SkillBrowser} addresses this problem through Meta-Skill-managed access, allowing agents to formulate capability-oriented queries and load selected Skills on demand.

\subsection{Selective Capability Use}

Beyond selecting which external capability is relevant, a complementary line of work studies whether external assistance should be used at all. In tool-augmented agents, When2Call explicitly evaluates decisions to invoke a tool, request clarification, or refrain from tool use, while When2Tool studies tool necessity under controlled capability boundaries and shows that agents frequently invoke tools even when their parametric capabilities suffice~\citep{when2call2025ross,when2tool2026sun}. Moving from tool invocation to retrieval, ProactAgent treats access to factual, episodic, and behavioral-skill experience as an explicit policy action and learns when and what to retrieve through process-level reinforcement learning~\citep{proactagent2026cai}. These works establish that capability access itself can be treated as a decision rather than mandatory.

Recent work has begun to study this question specifically for Agent Skills. SelSkill formulates Skill use as a decision-point-level \emph{skill-or-skip} problem and learns selective invocation through dual-granularity preference learning~\citep{selskill2026chen}. SkillApt instead separates retrieval from activation: after a candidate Skill has already been retrieved, it uses counterfactual evidence from matched executions to decide whether the Skill should actually be loaded~\citep{skillapt2026guo}. SkillGate trains in-policy, mid-episode Skill selection with action-local credit assignment over a visible candidate slate, reducing exposure to misleading Skills, although its selection-specific objective does not directly assign credit to abstention~\citep{skillgate2026li}. \textsc{SkillBrowser} differs in both the decision point and the mechanism: we study \emph{pre-retrieval Skill necessity}, allowing an agent to bypass Skill search entirely when external procedural guidance is unnecessary, while jointly formulating a capability-oriented query when search is warranted. Moreover, unlike learned invocation or activation policies, this control is realized through a training-free Meta-Skill and can be combined with different underlying Skill retrieval methods.

\section{Motivation}

Accessing a large Skill library requires deciding which Skills should enter the agent's working context. Existing paradigms make this decision imperfectly: progressive disclosure exposes metadata for a growing candidate catalog, while static retrieval exposes a small set of Skill contents selected from the initial task query. We formulate the resulting problem as \emph{Selective Skill Exposure}: a scalable Skill access mechanism should expose only the information necessary for the agent's current solving need, avoiding both irrelevant metadata and unnecessary procedural contents.

\subsection{Progressive Disclosure Does Not Eliminate Context Pollution}

As the Skill repository grows, progressive disclosure requires the agent to identify the needed capability from an increasingly large metadata catalog. We examine whether this growing metadata exposure impairs reliable Skill access.

\begin{figure}[t]
    \centering
    \includegraphics[width=\columnwidth]{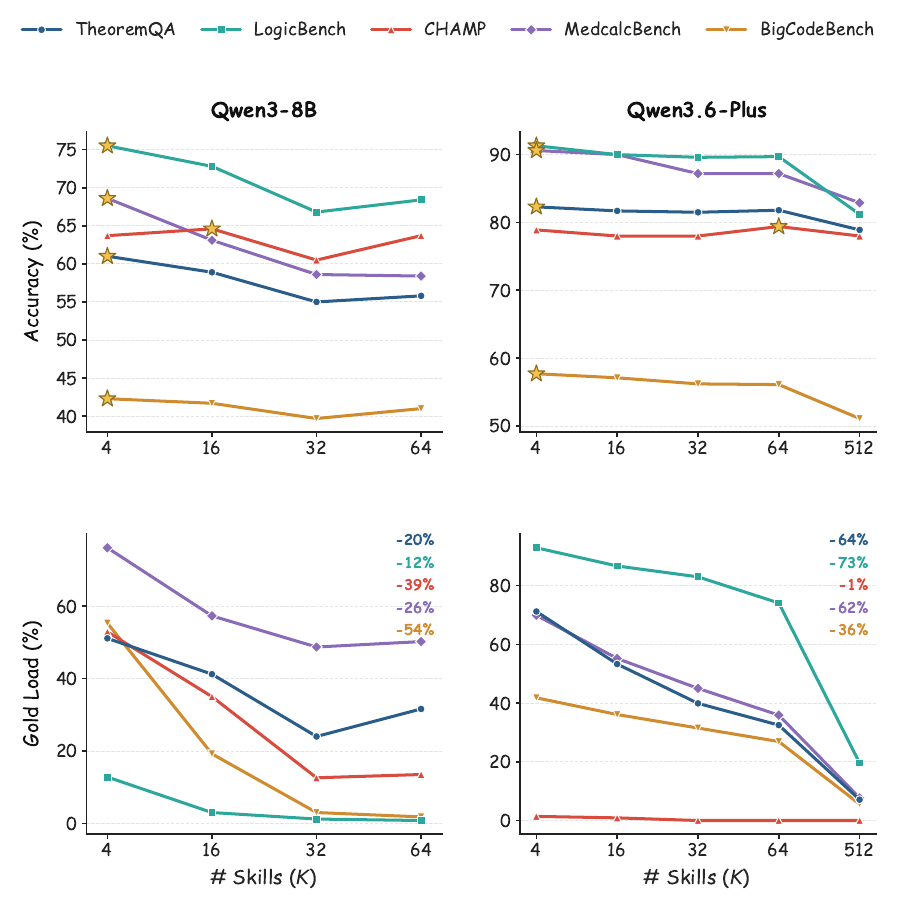}
    \caption{Effect of scaling the number of exposed skill descriptions under progressive disclosure.}
    \label{fig:fig2}
\end{figure}

We evaluate this effect by fixing the task instances, backbone model, and target Skill, while varying only the number of exposed Skill descriptions, denoted by $K$. Figure~\ref{fig:fig2} reports downstream accuracy and gold-Skill loading rate (Gold Load). Increasing $K$ consistently reduces the agent's ability to load the required Skill. For example, on \textsc{MedCalcBench}, Qwen3-8B's Gold Load decreases from $76.1$ at $K=4$ to $48.7$ at $K=32$, with accuracy dropping from $68.6$ to $58.6$. On \textsc{TheoremQA}, Qwen3.6-Plus's Gold Load further drops from $71.2$ at $K=4$ to $7.1$ at $K=512$.

Since only the number of exposed descriptions changes, this degradation reveals metadata-level context pollution: progressive disclosure reduces full-content exposure, but does not provide scalable selective access to large Skill repositories.

\subsection{Static Retrieval May Misalign with Agent Needs}

Static retrieval relies on the original task query to identify relevant Skills before solving. We examine whether this query adequately represents the procedural capability required by the agent.

To examine this assumption, we compare LLM gold-Skill identification with static retrieval under a fixed library size of $K=16$. Table~\ref{tab:query_retrieval_gap} reports Gold Load when models directly observe candidate Skill descriptions, together with Recall@$5$ for BM25 and SkillGrep static retrieval from the original query. Once candidates are available, both models identify the gold Skill with near-perfect reliability. In contrast, even SkillGrep retrieval reaches only $31.7$ Recall@$5$ on \textsc{LogicBench} and $32.1$ on \textsc{CHAMP}, far below the corresponding agent-side Gold Load rates.

This gap indicates that stronger matching alone is insufficient: the original query may describe task context or final objectives rather than the intermediate capability needed during solving. Thus, static retrieval bounds exposure size, but may expose Skills misaligned with the agent's current procedural need.

\begin{table}[t]
  \centering
  \small
  \setlength{\tabcolsep}{0.8pt}
  \renewcommand{\arraystretch}{1.30}
  \begin{tabular}{@{}lccccc@{}}
    \hline
    \textbf{Method}
    & \textbf{Theo.}
    & \textbf{Logic}
    & \textbf{CHAMP}
    & \textbf{MedCalc}
    & \textbf{BigCode} \\
    \hline
    Qwen3-8B GL
    & 99.3
    & 100.0
    & 98.7
    & 100.0
    & 99.1 \\
    Qwen3.6-Plus GL
    & 99.5
    & 98.5
    & 100.0
    & 100.0
    & 92.6 \\
    \hline
    BM25 R@$5$
    & 77.7
    & 28.7
    & 30.7
    & 66.3
    & 96.5 \\
    SkillGrep R@$5$
    & 82.0
    & 31.7
    & 32.1
    & 81.3
    & 86.9 \\
    \hline
  \end{tabular}
  \caption{\label{tab:query_retrieval_gap}
    Comparison between agent-side gold-skill identification and static retrieval from the original task query under a skill library of size $N=16$. \textbf{GL} denotes the gold-skill loading rate, and \textbf{R@$5$} denotes Recall@$5$.
  }
\end{table}

\subsection{Design Implication: Selective Skill Exposure}

Together, these observations show that scalable Skill access requires controlling how Skill contents enter the agent's context. Progressive disclosure forces the agent to select Skills from an increasingly polluted metadata context, while static retrieval may preselect Skills from an initial query that does not reflect the capability need emerging during solving. Furthermore, retrieved Skill contents should not be injected unconditionally when external support is unnecessary. We therefore formulate \emph{Selective Skill Exposure}: Skill contents should be introduced according to the agent's current capability need, rather than through global metadata exposure or one-shot query-based retrieval. We instantiate this principle with \textsc{SkillBrowser} in the next section.

\section{Methods}

\begin{figure*}[t]
    \centering
    \includegraphics[width=\textwidth]{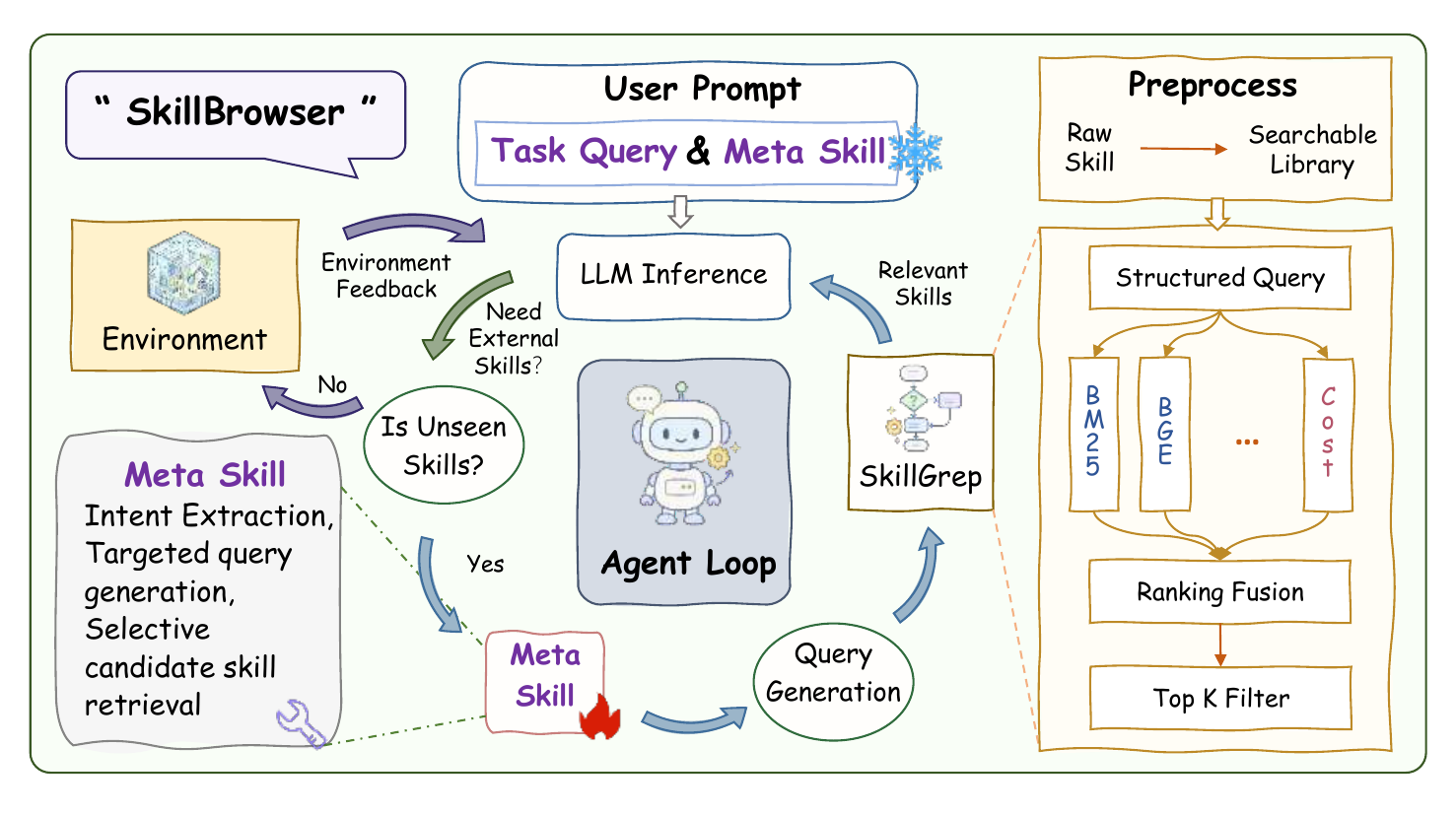}
    \caption{Overview of \textsc{SkillBrowser}. A reusable Meta-Skill enables the agent to decide whether external procedural support is needed and, when necessary, formulate a retrieval intent for \textsc{SkillGrep}. Only selected Skills are introduced into the agent--environment interaction loop.}
    \label{fig:pipeline}
\end{figure*}

We propose \textsc{SkillBrowser}, a Meta-Skill-managed framework for \emph{Selective Skill Exposure} over a large external Skill repository. As illustrated in Figure~\ref{fig:pipeline}, \textsc{SkillBrowser} separates scalable repository maintenance from online Skill access: \textsc{SkillGrep} maintains an offline searchable library, while a reusable Meta-Skill allows the solving agent to request and load external Skills only when needed.

\subsection{Selective Skill Exposure Formulation}

Let $\mathcal{S}=\{(m_j,f_j)\}_{j=1}^{N}$ denote an external Skill repository, where $m_j$ is the metadata of the $j$-th Skill and $f_j$ is its full procedural content. Let $P$ denote the user task and $c_{i-1}$ the solving context before the $i$-th potential Skill exposure step. For any subset $\mathcal{A}\subseteq\mathcal{S}$, we denote the corresponding full Skill contents by
\begin{equation}
    \mathcal{F}(\mathcal{A})
    =
    \{f_j \mid (m_j,f_j)\in\mathcal{A}\}.
\end{equation}

Following the definition of \emph{Selective Skill Exposure} introduced above, we formalize the ideal Skill exposure under the agent's current solving state. At step $i$, let
\begin{equation}
    z_i^\star \in \{0,1\}
\end{equation}
denote whether external Skill contents should be introduced, where $z_i^\star=0$ indicates that no additional Skill is needed and $z_i^\star=1$ indicates that external procedural support is required. When $z_i^\star=1$, let $\mathcal{S}_i^{\mathrm{need}}\subseteq\mathcal{S}$ denote the Skills providing the required capability. The ideal exposure is
\begin{equation}
    \mathcal{E}_i^\star
    =
    \begin{cases}
        \varnothing, & z_i^\star=0,\\
        \mathcal{F}(\mathcal{S}_i^{\mathrm{need}}), & z_i^\star=1.
    \end{cases}
\end{equation}

Existing paradigms instantiate exposure differently. Progressive disclosure conditions Skill selection on the exposed repository metadata:
\begin{equation}
    \mathcal{E}^{\mathrm{PD}}_{\mathrm{meta}}
    =
    \{m_j\}_{j=1}^{N}.
\end{equation}
Static retrieval instead directly exposes the contents of Skills retrieved from the initial task query:
\begin{equation}
    \mathcal{E}^{\mathrm{SR}}
    =
    \mathcal{F}\!\left(R(P,\mathcal{S},k)\right).
\end{equation}
Neither formulation represents the state-dependent ideal $\mathcal{E}_i^\star$: the former requires selection over repository metadata, while the latter fixes exposure through task-level retrieval without an explicit no-exposure decision.

Approximating $\mathcal{E}_i^\star$ requires exposing relevant Skills only when an external capability need arises during solving. \textsc{SkillBrowser} implements this by providing a Meta-Skill-managed search interface that the agent can invoke within its ordinary reasoning trajectory, while \textsc{SkillGrep} retrieves Skills matching the expressed need.

\subsection{Meta-Skill-Managed Agent Loop}

The online interaction process is shown in the center of Figure~\ref{fig:pipeline}. We provide the agent with a fixed Meta-Skill $s^\star$ together with the user task. Unlike task-specific Skills, $s^\star$ does not contain a procedure for solving a particular problem. Instead, it describes how the agent can access the external Skill repository when its ongoing reasoning requires additional procedural support.

Importantly, the Meta-Skill does not impose a separate retrieve-or-not decision stage. The agent proceeds with ordinary inference and environment interaction, and may invoke \textsc{SkillGrep} through the Meta-Skill whenever an external capability need arises. When such an invocation occurs at step $i$, the agent constructs a structured retrieval query
\begin{equation}
    q_i
    =
    Q_M(P,c_{i-1};s^\star),
\end{equation}
where $q_i$ expresses the capability needed under the current solving context rather than simply repeating the original task request.

In our implementation, $q_i$ may include a concise capability description, relevant task context, required or preferred capability attributes, type hints, positive or negative retrieval signals, and explicit constraints. These fields allow the agent to articulate diverse intermediate needs, refine retrieval when necessary, and reuse the same Meta-Skill across heterogeneous tasks.

Given the agent's invocation behavior, the actual repository-derived exposure introduced by \textsc{SkillBrowser} is
\begin{equation}
    \mathcal{E}_i^{\mathrm{SB}}
    =
    \begin{cases}
        \mathcal{F}\!\left(R(q_i,\mathcal{S},k)\right),
        & \text{invocation},\\
        \varnothing,
        & \text{otherwise.}
    \end{cases}
\end{equation}
Retrieved Skills are then returned to the agent loop as additional procedural context. Depending on the task, they may provide a formula, reasoning procedure, domain workflow, software usage pattern, or tool-use instruction, which the agent applies in subsequent inference and interaction.

This design approximates selective Skill exposure without prescribing when retrieval must occur: the Meta-Skill makes external Skill access available as a reusable capability, while the agent retains flexibility over whether and when to invoke it during problem solving. It relies on capability-need recognition but does not assume that this recognition is accurate. The ideal variable $z_i^\star$ is a conceptual target, not an observed supervision label; BER and UER diagnose the model's access behavior rather than certify correct self-assessment.

\subsection{SkillGrep: Capability-Oriented Skill Retrieval}

Once the agent invokes Skill search with a structured query $q_i$, \textsc{SkillGrep} retrieves a small set of Skills that approximate the capability required at the current solving state:
\begin{equation}
    \widehat{\mathcal{S}}_i
    =
    R(q_i,\mathcal{S},k)
    \approx
    \mathcal{S}_i^{\mathrm{need}}.
\end{equation}

As shown on the right side of Figure~\ref{fig:pipeline}, \textsc{SkillGrep} consists of offline repository construction and online retrieval. Offline, each raw Skill is converted into searchable representations derived from its metadata and instructional content, such as capability descriptions, usage conditions, representative queries, and the original instructions. This preprocessing provides a shared retrieval interface for heterogeneous Skills while preserving their original contents.

Online, \textsc{SkillGrep} matches the agent-generated query against these representations using hybrid retrieval. It combines lexical matching, which captures precise terms and identifiers, with semantic matching, which captures paraphrased capability needs, and fuses the resulting rankings to return the top-$k$ Skills. The contents of the retrieved Skills are then returned to the agent loop for subsequent application. Further implementation details are provided in Appendix~\ref{app:SkillGrep}.

\section{Experiments}
\label{sec:experiments}

\subsection{Experimental Setup}

\paragraph{Benchmarks.}
We evaluate \textsc{SkillBrowser} on six subsets of SRA-Bench~\citep{SRA2026su}: \textsc{TheoremQA}, \textsc{LogicBench}, \textsc{CHAMP}, \textsc{MedCalcBench}, \textsc{BigCodeBench}, and \textsc{ToolQA}. The first five cover theorem application, logical reasoning, mathematical problem solving, clinical calculation, and software-library usage. ToolQA additionally evaluates multi-step tool interaction, jointly testing Skill retrieval, exposure, tool selection, and execution. All full-corpus retrieval experiments use the same fixed repository of 26,262 candidate Skills. Appendix~\ref{app:benchmarks} gives benchmark details.

\paragraph{Models.}
We use five LLMs from different model families: Qwen3-8B, Qwen3.6-Plus, GLM-5.1, GPT-5.4-mini, and Gemini-3.1-Flash-Lite.

\paragraph{Methods.}
\textbf{SkillBrowser} combines Meta-Skill-managed access with \textsc{SkillGrep}. \textbf{Static BM25} and \textbf{Static SkillGrep} retrieve from the initial task query and expose the retrieved contents before solving; Static SkillGrep shares the full framework's retrieval backend. \textbf{LLM Selection} follows the SR-Agent baseline~\citep{SRA2026su}, selecting Skills from a precomputed BM25 top-50 candidate set. \textbf{Gold Exposure} directly provides the annotated gold Skill as a relevance-controlled reference, not a performance upper bound. \textbf{Meta-Skill+BM25} retains managed access with BM25 retrieval for component analysis. Progressive disclosure is evaluated separately as a feasible-budget scaling diagnostic (Figure~\ref{fig:fig2}), since exposing metadata for all 26,262 Skills exceeds the evaluated context budget.

\paragraph{Retrieval configuration.}
Dense retrieval uses \texttt{bge-base-en-v1.5} with mean pooling and L2 normalization. Retrieval evaluation uses $k=5$. Multi-view weights, fusion weights, and the abstention threshold were selected by grid search on the development set; Appendix~\ref{app:SkillGrep} specifies the available scoring configuration.

\paragraph{Metrics.}
We report end-task accuracy as the primary metric. For Selective Skill Exposure, we additionally analyze two subsets constructed separately for each model--benchmark pair: \emph{skill-beneficial} instances, which are solved incorrectly without Skills but correctly with the gold Skill, and \emph{skill-unnecessary} instances, which are solved correctly under both conditions. We report Beneficial Exposure Rate (BER) on the former and Unnecessary Exposure Rate (UER) on the latter, each defined as the fraction of instances receiving retrieved Skill contents. These are outcome-conditioned, model-dependent diagnostic partitions, not intrinsic labels of Skill necessity; they omit instances incorrect under both conditions and those harmed by gold exposure. BER measures exposure to any retrieved Skill, not successful gold-Skill retrieval. We additionally report retrieval quality, end-to-end resource use, and paired correctness comparisons.

\begin{table*}[t]
\centering
\footnotesize
\setlength{\tabcolsep}{4pt}
\renewcommand{\arraystretch}{1.08}
\begin{tabular}{@{}llrrrrrr@{}}
\toprule
\textbf{Model} & \textbf{Method} & \textsc{Theo.} & \textsc{Logic} & \textsc{CHAMP} & \textsc{MedCalc} & \textsc{BigCode} & \textsc{ToolQA} \\
\midrule
\multirow{5}{*}{Qwen3-8B} & \textbf{SkillBrowser} & 71.0 & 79.5 & 68.2 & 65.4 & 46.7 & 44.0 \\
 & Static \textsc{SkillGrep} & 61.7 & 72.0 & 69.0 & 60.6 & 46.7 & 17.0 \\
 & Static BM25 & 53.0 & 72.0 & 67.3 & 51.4 & 45.8 & 17.0 \\
 & LLM Selection & 56.0 & 80.0 & 61.0 & 71.0 & 42.0 & 13.0 \\
 & \cellcolor{refgray}Gold Exposure & \cellcolor{refgray}58.9 & \cellcolor{refgray}87.0 & \cellcolor{refgray}66.8 & \cellcolor{refgray}75.0 & \cellcolor{refgray}50.0 & \cellcolor{refgray}51.0 \\
\midrule
\multirow{5}{*}{Qwen3.6-Plus} & \textbf{SkillBrowser} & 84.5 & 84.1 & 81.2 & 93.0 & 61.0 & 57.0 \\
 & Static \textsc{SkillGrep} & 83.1 & 80.9 & 82.1 & 90.1 & 57.3 & 27.0 \\
 & Static BM25 & 83.9 & 80.0 & 81.6 & 87.8 & 58.3 & 32.0 \\
 & LLM Selection & 83.0 & 84.0 & 80.0 & 92.0 & 54.0 & 40.0 \\
 & \cellcolor{refgray}Gold Exposure & \cellcolor{refgray}83.5 & \cellcolor{refgray}90.5 & \cellcolor{refgray}84.3 & \cellcolor{refgray}91.7 & \cellcolor{refgray}59.7 & \cellcolor{refgray}76.0 \\
\midrule
\multirow{5}{*}{GLM-5.1} & \textbf{SkillBrowser} & 82.5 & 84.3 & 74.9 & 90.7 & 61.7 & 51.0 \\
 & Static \textsc{SkillGrep} & 78.1 & 80.3 & 79.4 & 80.5 & 59.3 & 39.0 \\
 & Static BM25 & 77.9 & 73.0 & 78.5 & 73.2 & 60.0 & 37.0 \\
 & LLM Selection & 78.0 & 84.0 & 77.0 & 88.0 & 57.0 & 41.0 \\
 & \cellcolor{refgray}Gold Exposure & \cellcolor{refgray}78.1 & \cellcolor{refgray}89.7 & \cellcolor{refgray}80.3 & \cellcolor{refgray}87.2 & \cellcolor{refgray}60.0 & \cellcolor{refgray}73.0 \\
\midrule
\multirow{5}{*}{\makecell[l]{GPT-5.4-\\mini}} & \textbf{SkillBrowser} & 80.7 & 71.7 & 78.9 & 85.0 & 62.0 & 43.0 \\
 & Static \textsc{SkillGrep} & 76.7 & 77.7 & 76.2 & 81.0 & 57.7 & 32.0 \\
 & Static BM25 & 74.3 & 68.3 & 73.5 & 81.0 & 59.3 & 33.0 \\
 & LLM Selection & 75.0 & 81.0 & 78.0 & 83.0 & 55.0 & 31.0 \\
 & \cellcolor{refgray}Gold Exposure & \cellcolor{refgray}78.0 & \cellcolor{refgray}85.3 & \cellcolor{refgray}76.2 & \cellcolor{refgray}84.1 & \cellcolor{refgray}63.0 & \cellcolor{refgray}71.0 \\
\midrule
\multirow{5}{*}{\makecell[l]{Gemini-3.1-\\Flash-Lite}} & \textbf{SkillBrowser} & 81.3 & 69.3 & 74.9 & 83.0 & 58.0 & 61.0 \\
 & Static \textsc{SkillGrep} & 81.0 & 80.7 & 75.3 & 82.6 & 56.3 & 18.0 \\
 & Static BM25 & 83.0 & 72.7 & 69.1 & 79.9 & 57.7 & 17.0 \\
 & LLM Selection & 81.0 & 75.0 & 79.0 & 82.0 & 52.0 & 38.0 \\
 & \cellcolor{refgray}Gold Exposure & \cellcolor{refgray}83.0 & \cellcolor{refgray}87.7 & \cellcolor{refgray}71.3 & \cellcolor{refgray}81.7 & \cellcolor{refgray}59.3 & \cellcolor{refgray}70.0 \\
\bottomrule
\end{tabular}
\caption{End-task accuracy (\%). Static \textsc{SkillGrep} shares the retrieval backend of \textsc{SkillBrowser}. LLM Selection selects from a BM25 top-50 set. Gray rows denote Gold Exposure, a relevance-controlled reference rather than an upper bound. }
\label{tab:main_results}
\end{table*}

\subsection{Main Results}

Table~\ref{tab:main_results} reports accuracy across five models and six benchmarks. \textsc{SkillBrowser} outperforms Static BM25 in 26 of 30 settings and Static \textsc{SkillGrep} in 23, with one tie against the latter. Because Static SkillGrep uses the same retrieval backend, this comparison controls for backend strength while changing both query formulation and exposure timing. It does not isolate the no-exposure decision from capability-oriented query generation.

Against LLM Selection, the reported comparison yields 21 wins, four ties, and five losses at whole-percentage-point reporting precision. Improvements are task-dependent: the advantage over Static SkillGrep is consistent across all five models on \textsc{TheoremQA}, \textsc{MedCalcBench}, and \textsc{ToolQA}, whereas losses remain on \textsc{CHAMP} and \textsc{LogicBench}. On ToolQA, SkillBrowser reaches $43$--$61\%$ accuracy versus $17$--$39\%$ for static retrieval. Appendix~\ref{app:toolqa} details retrieval behavior.

Gold Exposure does not consistently achieve the highest accuracy: SkillBrowser surpasses it in 13 of 30 settings, comprising four on TheoremQA, four on MedCalcBench, three on CHAMP, and two on BigCodeBench. This does not establish that retrieved Skills are more relevant than the gold annotations. Relevant instructions may be redundant, may have mismatched granularity, or may alter an otherwise successful reasoning trajectory. The transition analysis in Appendix~\ref{app:gold_transitions} examines both gains and failures without attributing all improvements to skipping retrieval.

\paragraph{Paired statistical evidence.}
On valid paired predictions pooled across six datasets, SkillBrowser improves over Static SkillGrep ($78.59\%$ vs.\ $74.64\%$; $N=11{,}736$; two-sided exact McNemar $p=1.46\times10^{-28}$) and LLM Selection ($75.93\%$ vs.\ $72.82\%$; $p=4.41\times10^{-10}$), while remaining below Gold Exposure ($78.59\%$ vs.\ $80.52\%$; $p=1.49\times10^{-8}$). These pooled comparisons use baseline-specific valid intersections, so their accuracies need not match macro-averages of Table~\ref{tab:main_results}. They support aggregate differences, not uniform gains or significance in every setting; Appendix~\ref{app:statistics} reports discordant counts and interpretation limits.

\begin{table}[t]
\centering
\small
\setlength{\tabcolsep}{3.2pt}
\renewcommand{\arraystretch}{1.15}
\begin{tabular}{@{}lrrrrr@{}}
\toprule
& \multicolumn{2}{c}{\textbf{Exposure} (\%)}
& \multicolumn{2}{c}{\textbf{Accuracy} (\%)} \\
\cmidrule(lr){2-3}\cmidrule(l){4-5}
\textbf{Model} & \textbf{SB.} & \textbf{SG.}
& \textbf{SB.} & \textbf{SG.} & \textbf{$N$} \\
\midrule
\multicolumn{6}{l}{\textit{Skill-Beneficial} ($\uparrow$ exposure; $\uparrow$ accuracy)} \\
Qwen3-8B       & 74.9 & 100.0 & \best{67.1} & \best{67.1} & 859 \\
Qwen3.6-Plus   & 80.3 & 100.0 & 76.2 & \best{77.4} & 336 \\
GLM-5.1        & 84.0 & 100.0 & \best{85.3} & 77.1 & 556 \\
GPT-5.4-mini   & 30.3 & 100.0 & 46.2 & \best{72.4} & 145 \\
\makecell[l]{Gemini-3.1-\\Flash-Lite} & 34.0 & 100.0 & 26.2 & \best{74.5} & 141 \\
\midrule
\multicolumn{6}{l}{\textit{Skill-Unnecessary} ($\downarrow$ exposure; $\uparrow$ accuracy)} \\
Qwen3-8B       & 66.5 & 100.0 & \best{95.2} & 85.4 & 1216 \\
Qwen3.6-Plus   & 77.4 & 100.0 & \best{99.1} & 95.8 & 2173 \\
GLM-5.1        & 88.2 & 100.0 & \best{98.2} & 95.1 & 1852 \\
GPT-5.4-mini   & 45.1 & 100.0 & \best{98.8} & 93.6 & 763 \\
\makecell[l]{Gemini-3.1-\\Flash-Lite} & 43.4 & 100.0 & \best{97.4} & 93.7 & 772 \\
\bottomrule
\end{tabular}
\caption{
Selective Skill Exposure analysis aggregated across the four diagnostic benchmarks (TheoremQA, LogicBench, CHAMP, and MedCalcBench).
\textbf{SB.} denotes the full \textsc{SkillBrowser} pipeline and
\textbf{SG.} denotes static \textsc{SkillGrep} retrieval.
Blue-shaded bold values indicate the better accuracy between the two methods.
Full model--benchmark results are provided in Appendix~\ref{app:selective_exposure}.
}
\label{tab:selective_exposure}
\end{table}

\subsection{Selective Skill Exposure Analysis}
\label{sec:selective_exposure}

Table~\ref{tab:selective_exposure} compares Meta-Skill-managed access with Static \textsc{SkillGrep} under the same retrieval backend. Static \textsc{SkillGrep} exposes retrieved Skill contents for every query, whereas Meta-Skill-managed access allows the agent to decide whether external procedural support should enter its reasoning context. This comparison controls for retrieval-backend strength; its effect jointly reflects the decision to search and the capability-oriented queries generated when searching.

\paragraph{Avoiding unnecessary exposure.}
On skill-unnecessary instances, Meta-Skill-managed access reduces exposure and improves accuracy for every model after aggregation across the four diagnostic benchmarks. Since these instances are already solvable without external Skills, the result shows that even retrieved procedural contents can disturb otherwise correct reasoning when introduced unconditionally. This finding directly supports SSE: Skill access requires controlling not only which Skills are retrieved, but also whether their contents should be exposed to the agent at all.

\paragraph{Selective access to beneficial Skills.}
The fine-grained results in Appendix~\ref{app:selective_exposure} reveal model-dependent behavior. Qwen3-8B, Qwen3.6-Plus, and GLM-5.1 expose Skills more often on beneficial than unnecessary instances in multiple benchmarks. However, GPT-5.4-mini and Gemini-3.1-Flash-Lite frequently under-invoke Skills on beneficial instances, particularly on LogicBench and CHAMP. The framework makes capability-need recognition explicit but does not solve it. These outcome-conditioned subsets diagnose such failures for a particular backbone and run; they do not establish model-independent Skill necessity or a causal effect of exposure alone.

\begin{table}[t]
\centering
\footnotesize
\setlength{\tabcolsep}{2pt}
\renewcommand{\arraystretch}{1.10}
\begin{tabular}{@{}llcccc@{}}
\toprule
\textbf{Exposure} & \textbf{Retriever}
& \textsc{Theo.}
& \textsc{Logic}
& \textsc{CHAMP}
& \textsc{MedCalc}\\
\midrule

Static
& BM25
& 74.42 & 73.20 & 74.00 & 74.66\\

Static
& \textsc{SG.}
& 76.12 & \cellcolor{bestblue}\textbf{78.32} & \cellcolor{bestblue}\textbf{76.40} & 78.96\\

Meta
& BM25
& 75.52 & 74.32 & 71.40 & 77.66 \\

Meta
& \textsc{SG.}
& \cellcolor{bestblue}\textbf{80.00} & 77.78 & 75.62 & \cellcolor{bestblue}\textbf{83.42}\\
\bottomrule
\end{tabular}
\caption{
Component analysis averaged across backbone models, measured by end-task accuracy (\%).
Static and Meta denote static exposure and Meta-Skill-managed exposure, respectively;
\textsc{SG.} denotes \textsc{SkillGrep}.
Meta + \textsc{SG.} is the full \textsc{SkillBrowser} framework.
Blue-shaded values indicate the best result in each column.
}
\label{tab:component_ablation}
\end{table}

\subsection{Component Analysis}

Table~\ref{tab:component_ablation} examines how the retrieval backend and exposure policy contribute to the effectiveness of \textsc{SkillBrowser}.

\paragraph{Effect of \textsc{SkillGrep}.}
Under static exposure, replacing BM25 with \textsc{SkillGrep} improves average accuracy by $3.38$ points. The largest gains occur on \textsc{LogicBench} and \textsc{MedCalcBench} ($+5.12$ and $+4.30$ points), showing that retrieval-oriented Skill representations and hybrid matching improve the quality of Skills selected from the original query.

\paragraph{Effect of Meta-Skill-managed exposure.}
With BM25 retrieval fixed, Meta-Skill-managed exposure yields a modest average improvement, with its clearest benefit on \textsc{MedCalcBench} ($+3.00$ points). This is consistent with its long clinical queries, where the required procedure can be better articulated during solving. In contrast, performance decreases on \textsc{CHAMP} ($-2.60$ points), where models may not request external support for tasks that appear solvable through internal reasoning.

\paragraph{Complementarity of the two components.}
Combining Meta-Skill-managed exposure with \textsc{SkillGrep} produces the strongest overall configuration, improving over Static + \textsc{SkillGrep} by $1.76$ points on average. The gains are concentrated on \textsc{TheoremQA} and \textsc{MedCalcBench} ($+3.88$ and $+4.46$ points), while small drops remain on \textsc{LogicBench} and \textsc{CHAMP}. Together with the SSE analysis, these results indicate that improved retrieval and selective exposure are complementary when the required procedural capability is identifiable during solving, while recognizing beneficial Skill use remains challenging for tasks resembling ordinary internal reasoning.

\paragraph{Retrieval components and stronger retrievers.}
Using initial task descriptions as queries over the same 26,262-Skill corpus, SkillGrep leads the SRA-Bench BGE, TF--IDF, BM25, and Contriever baselines on all six reported retrieval metrics (Appendix~\ref{app:retrieval_evaluation}). Leave-one-component-out ablations show contributions from dense retrieval, sparse views, reciprocal-rank-fusion scoring, and capability/type signals. GoldHit@$5$ drops from $0.623$ to $0.463$ without dense retrieval and to $0.462$ without capability/type signals. Removing the compatibility penalty has negligible effect, so it is not a demonstrated source of gains on this benchmark.

\subsection{Efficiency and Access Overhead}
Across five models with 100 valid ToolQA trajectories each, SkillBrowser makes 383 retrieval calls on 500 trajectories ($0.766$ calls per trajectory). Every searching trajectory invokes SkillGrep exactly once and loads five candidates, yielding $3.83$ loaded Skills per trajectory; $278/383$ calls retrieve a gold Skill ($72.6\%$). These are cross-model totals, distinct from the GLM-5.1 token and latency measurements.

Total ToolQA LLM calls vary by backbone: SkillBrowser uses more calls than LLM Selection for four models but fewer for GLM-5.1. For GLM-5.1, combined input/output token use is lower on three of six benchmarks and higher on the other three, whereas measured latency is higher on all six. LLM Selection uses precomputed BM25 candidates and its latency excludes that retrieval stage. Thus, selective access does not imply uniformly lower cost or latency. Appendix~\ref{app:efficiency} reports complete-trajectory token counts, LLM calls, and measured latency.

\section{Conclusion}

In this work, we formulate \emph{Selective Skill Exposure} as a central problem for scaling Skill-using LLM agents: procedural support should be introduced according to the agent's current need, without exposing excessive metadata or unnecessary Skill contents. We show that existing paradigms fall short in complementary ways. Progressive disclosure remains vulnerable to metadata-level context pollution, while static retrieval may expose Skills selected from an initial query misaligned with the capability needed during reasoning.

We propose \textsc{SkillBrowser}, a Meta-Skill-managed framework for SSE. Rather than selecting Skills from a globally exposed catalog or unconditionally injecting retrieved contents, it makes Skill exposure an agent-initiated action: when procedural support is needed, the agent formulates a capability-oriented query through the Meta-Skill and retrieves relevant Skills from \textsc{SkillGrep} for on-demand use.

Experiments on SRA-Bench show that metadata exposure impairs Skill loading at scale, while \textsc{SkillBrowser} improves end-task accuracy over static query-based retrieval, especially when the required procedure is obscured by noisy context. Meta-Skill-managed access further mitigates the degradation caused by unnecessary Skill exposure on already-solvable instances. The added ToolQA evaluation extends this evidence to multi-step tool interaction, while comparisons with LLM Selection support aggregate gains over a stronger deployable baseline. Under-invocation, task-dependent losses, and additional latency remain limitations of the approach.

\section*{Limitations}

\textsc{SkillBrowser} depends on imperfect capability-need recognition. Models may under-invoke useful Skills on problems resembling ordinary internal reasoning, especially LogicBench and CHAMP. The beneficial/unnecessary partitions are model- and outcome-dependent diagnostics, not intrinsic necessity labels. They omit two reference-outcome groups and do not capture variability across repeated runs. Improving need recognition through targeted training, preference optimization, or uncertainty modeling is future work.

Our six-subset evaluation includes multi-step ToolQA interactions but uses a fixed repository. It does not test continual Skill addition, removal, or updating, persistent cross-session reuse, or long-horizon interaction with an evolving repository. A static corpus isolates exposure behavior from indexing, versioning, and invalidation. The separation between the agent interface and offline index could support dynamic repositories, but this extensibility has not been evaluated.

Retrieval adds latency, and token savings depend on the task. The LLM Selection timing excludes precomputed BM25 retrieval, so reported latency is not a fully symmetric online-system comparison. Pooled significance tests do not establish gains for every model or dataset; repeated task instances across backbones can also induce dependence among pooled observations. Broader sensitivity studies of $k$, encoder choice, and abstention thresholds remain open.

Real repositories may contain erroneous, incomplete, or unsafe instructions. SSE addresses when and which Skills enter context; it does not verify their trustworthiness or guarantee safe execution. Provenance review at ingestion, content validation before loading, permission checks before tool use, and execution sandboxing are possible deployment safeguards, none of which is established by the present evaluation.
\bibliography{custom}

\begin{thebibliography}{25}
\providecommand{\natexlab}[1]{#1}

\bibitem[{Braunschweiler et~al.(2025)Braunschweiler, Doddipatla, and
  Zorila}]{ToolReAGt2025braunschweiler}
Norbert Braunschweiler, Rama Doddipatla, and Tudor-catalin Zorila. 2025.
\newblock \href {https://doi.org/10.18653/v1/2025.knowllm-1.7}
  {{T}ool{R}e{AG}t: Tool retrieval for {LLM}-based complex task solution via
  retrieval augmented generation}.
\newblock In \emph{Proceedings of the 3rd Workshop on Towards Knowledgeable
  Foundation Models (KnowFM)}, pages 75--83, Vienna, Austria. Association for
  Computational Linguistics.

\bibitem[{Cai et~al.(2026)Cai, Li, Zhou, Chen, Li, Zhang, and
  He}]{proactagent2026cai}
Yuxuan Cai, Wei Li, Jie Zhou, Qin Chen, Xin Li, Bo~Zhang, and Liang He. 2026.
\newblock \href {https://arxiv.org/abs/2604.20572} {Ask only when needed:
  Proactive retrieval from memory and skills for experience-driven lifelong
  agents}.
\newblock \emph{Preprint}, arXiv:2604.20572.

\bibitem[{Chen et~al.(2026)Chen, Lin, Sun, Yang, Wang, Qin, Hu, Pan, and
  Zeng}]{selskill2026chen}
Chishui Chen, Jiaye Lin, Te~Sun, Yi~Yang, Junxi Wang, Cong Qin, Yangen Hu,
  Lu~Pan, and Ke~Zeng. 2026.
\newblock \href {https://arxiv.org/abs/2606.00510} {Skill or skip? learning
  selective skill invocation in agentic tasks via dual-granularity preference
  learning}.
\newblock \emph{Preprint}, arXiv:2606.00510.

\bibitem[{Cho et~al.(2026)Cho, Kang, and Kim}]{hongcheol_2026}
Hongcheol Cho, Ryangkyung Kang, and Youngeun Kim. 2026.
\newblock \href {https://doi.org/10.48550/ARXIV.2605.05726} {Skillret: A
  large-scale benchmark for skill retrieval in llm agents}.

\bibitem[{Fei et~al.(2025)Fei, Zheng, and
  Feng}]{fei2025mcpzeroactivetooldiscovery}
Xiang Fei, Xiawu Zheng, and Hao Feng. 2025.
\newblock \href {https://arxiv.org/abs/2506.01056} {Mcp-zero: Active tool
  discovery for autonomous llm agents}.
\newblock \emph{Preprint}, arXiv:2506.01056.

\bibitem[{Guo et~al.(2026)Guo, Sui, Jiang, Li, Li, Li, and
  Ou}]{skillapt2026guo}
Shuang Guo, Xiaobo Sui, Fei Jiang, Ruixin Li, Manxiang Li, Rong Li, and Ziyang
  Ou. 2026.
\newblock \href {https://arxiv.org/abs/2609.26863} {Towards strategy-level rsi
  for skill-augmented agents: Learning when to reuse skills from execution
  feedback}.
\newblock \emph{Preprint}, arXiv:2609.26863.

\bibitem[{Han et~al.(2026)Han, Zhang, Song, Fang, Chen, Sun, and
  Hu}]{han2026sweskillsbenchagentskillsactually}
Tingxu Han, Yi~Zhang, Wei Song, Chunrong Fang, Zhenyu Chen, Youcheng Sun, and
  Lijie Hu. 2026.
\newblock \href {https://arxiv.org/abs/2603.15401} {Swe-skills-bench: Do agent
  skills actually help in real-world software engineering?}
\newblock \emph{Preprint}, arXiv:2603.15401.

\bibitem[{Jiang et~al.(2026)Jiang, Li, Deng, Ma, Wang, Wang, and
  Yu}]{sokagenticskills2026jiang}
Yanna Jiang, Delong Li, Haiyu Deng, Baihe Ma, Xu~Wang, Qin Wang, and Guangsheng
  Yu. 2026.
\newblock \href {https://arxiv.org/abs/2602.20867} {Sok: Agentic skills --
  beyond tool use in llm agents}.
\newblock \emph{Preprint}, arXiv:2602.20867.

\bibitem[{Li et~al.(2023)Li, Zhao, Yu, Song, Li, Yu, Li, Huang, and
  Li}]{APIBank2023li}
Minghao Li, Yingxiu Zhao, Bowen Yu, Feifan Song, Hangyu Li, Haiyang Yu, Zhoujun
  Li, Fei Huang, and Yongbin Li. 2023.
\newblock \href {https://doi.org/10.18653/v1/2023.emnlp-main.187} {{API}-bank:
  A comprehensive benchmark for tool-augmented {LLM}s}.
\newblock In \emph{Proceedings of the 2023 Conference on Empirical Methods in
  Natural Language Processing}, pages 3102--3116, Singapore. Association for
  Computational Linguistics.

\bibitem[{Li et~al.(2026{\natexlab{a}})Li, Jiao, Shao, Zhang, Lu, Guo, Liu,
  Zhang, and Yu}]{skillgate2026li}
Qingyao Li, Wenxiang Jiao, Shuai Shao, Kangning Zhang, Yuan Lu, Yi~Guo, Weiwen
  Liu, Weinan Zhang, and Yong Yu. 2026{\natexlab{a}}.
\newblock \href {https://arxiv.org/abs/2608.18852} {Skillgate: Training
  in-policy skill selection in long-horizon agents}.
\newblock \emph{Preprint}, arXiv:2608.18852.

\bibitem[{Li et~al.(2026{\natexlab{b}})Li, Chen, Liu, Zheng, Chen, He, Li, You,
  Shen, Sun, Wang, Li, Zeng, Wang, Zhao, Wang, Chaim, Di, Gao, He, He, Jing,
  Kong, Lan, Li, Li, Li, Lin, Liu, Liu, Lyu, Ma, Wang, Wang, Wang, Ye, Zhang,
  Xing, Xue, Dillmann, and chung
  Lee}]{li2026skillsbenchbenchmarkingagentskills}
Xiangyi Li, Wenbo Chen, Yimin Liu, Shenghan Zheng, Xiaokun Chen, Yifeng He,
  Yubo Li, Bingran You, Haotian Shen, Jiankai Sun, Shuyi Wang, Binxu Li,
  Qunhong Zeng, Di~Wang, Xuandong Zhao, Yuanli Wang, Roey~Ben Chaim, Zonglin
  Di, Yipeng Gao, and 22 others. 2026{\natexlab{b}}.
\newblock \href {https://arxiv.org/abs/2602.12670} {Skillsbench: Benchmarking
  how well agent skills work across diverse tasks}.
\newblock \emph{Preprint}, arXiv:2602.12670.

\bibitem[{Liu et~al.(2026)Liu, Li, Du, Wu, Gui, Kuang, and
  Sun}]{liu2026graphskillsdependencyawarestructural}
Dawei Liu, Zongxia Li, Hongyang Du, Xiyang Wu, Shihang Gui, Yongbei Kuang, and
  Lichao Sun. 2026.
\newblock \href {https://arxiv.org/abs/2604.05333} {Graph of skills:
  Dependency-aware structural retrieval for massive agent skills}.
\newblock \emph{Preprint}, arXiv:2604.05333.

\bibitem[{Lumer et~al.(2025)Lumer, Gulati, Subbiah, Basavaraju, and
  Burke}]{lumer2025scalemcpdynamicautosynchronizingmodel}
Elias Lumer, Anmol Gulati, Vamse~Kumar Subbiah, Pradeep~Honaganahalli
  Basavaraju, and James~A. Burke. 2025.
\newblock \href {https://arxiv.org/abs/2505.06416} {Scalemcp: Dynamic and
  auto-synchronizing model context protocol tools for llm agents}.
\newblock \emph{Preprint}, arXiv:2505.06416.

\bibitem[{Patel et~al.(2026)Patel, Belli, Jalalirad, Arnold, Ermolov, and
  Major}]{patel2026dynamictooldependencyretrieval}
Bhrij Patel, Davide Belli, Amir Jalalirad, Maximilian Arnold, Aleksandr
  Ermolov, and Bence Major. 2026.
\newblock \href {https://arxiv.org/abs/2512.17052} {Dynamic tool dependency
  retrieval for lightweight function calling}.
\newblock \emph{Preprint}, arXiv:2512.17052.

\bibitem[{Patil et~al.(2024)Patil, Zhang, Wang, and
  Gonzalez}]{Gorilla2024patil}
Shishir~G. Patil, Tianjun Zhang, Xin Wang, and Joseph~E. Gonzalez. 2024.
\newblock \href {https://doi.org/10.52202/079017-4020} {Gorilla: Large language
  model connected with massive apis}.
\newblock In \emph{Advances in Neural Information Processing Systems},
  volume~37, pages 126544--126565. Curran Associates, Inc.

\bibitem[{Qin et~al.(2024)Qin, Liang, Ye, Zhu, Yan, Lu, Lin, Cong, Tang, Qian,
  Zhao, Hong, Tian, Xie, Zhou, Gerstein, li, Liu, and Sun}]{ToolLLM2024qin}
Yujia Qin, Shihao Liang, Yining Ye, Kunlun Zhu, Lan Yan, Yaxi Lu, Yankai Lin,
  Xin Cong, Xiangru Tang, Bill Qian, Sihan Zhao, Lauren Hong, Runchu Tian,
  Ruobing Xie, Jie Zhou, Mark Gerstein, dahai li, Zhiyuan Liu, and Maosong Sun.
  2024.
\newblock \href
  {https://proceedings.iclr.cc/paper_files/paper/2024/file/28e50ee5b72e90b50e7196fde8ea260e-Paper-Conference.pdf}
  {Toolllm: Facilitating large language models to master 16000+ real-world
  apis}.
\newblock In \emph{International Conference on Learning Representations},
  volume 2024, pages 9695--9717.

\bibitem[{Ross et~al.(2025)Ross, Mahabaleshwarkar, and
  Suhara}]{when2call2025ross}
Hayley Ross, Ameya~Sunil Mahabaleshwarkar, and Yoshi Suhara. 2025.
\newblock \href {https://doi.org/10.18653/v1/2025.naacl-long.174}
  {{W}hen2{C}all: When (not) to call tools}.
\newblock In \emph{Proceedings of the 2025 Conference of the Nations of the
  Americas Chapter of the Association for Computational Linguistics: Human
  Language Technologies (Volume 1: Long Papers)}, pages 3391--3409,
  Albuquerque, New Mexico. Association for Computational Linguistics.

\bibitem[{Shi et~al.(2025)Shi, Wang, Yan, Ren, Wang, Yin, and
  Ren}]{ToolRet2025shi}
Zhengliang Shi, Yuhan Wang, Lingyong Yan, Pengjie Ren, Shuaiqiang Wang, Dawei
  Yin, and Zhaochun Ren. 2025.
\newblock \href {https://doi.org/10.18653/v1/2025.findings-acl.1258} {Retrieval
  models aren{'}t tool-savvy: Benchmarking tool retrieval for large language
  models}.
\newblock In \emph{Findings of the Association for Computational Linguistics:
  ACL 2025}, pages 24497--24524, Vienna, Austria. Association for Computational
  Linguistics.

\bibitem[{Su et~al.(2026)Su, Long, Ai, Tang, Wang, Tu, and Liu}]{SRA2026su}
Weihang Su, Jianming Long, Qingyao Ai, Yichen Tang, Changyue Wang, Yiteng Tu,
  and Yiqun Liu. 2026.
\newblock \href {https://arxiv.org/abs/2604.24594} {Skill retrieval
  augmentation for agentic ai}.
\newblock \emph{Preprint}, arXiv:2604.24594.

\bibitem[{Sun et~al.(2026)Sun, Liu, Yan, Wang, and Weng}]{when2tool2026sun}
Chung-En Sun, Linbo Liu, Ge~Yan, Zimo Wang, and Tsui-Wei Weng. 2026.
\newblock \href {https://arxiv.org/abs/2605.09252} {Llm agents already know
  when to call tools -- even without reasoning}.
\newblock \emph{Preprint}, arXiv:2605.09252.

\bibitem[{Wang et~al.(2025)Wang, Han, Ji, Wang, Baldwin, and
  Li}]{ToolGen2025wang}
Renxi Wang, Xudong Han, Lei Ji, Shu Wang, Timothy Baldwin, and Haonan Li. 2025.
\newblock \href {https://openreview.net/forum?id=XLMAMmowdY} {Toolgen: Unified
  tool retrieval and calling via generation}.
\newblock In \emph{The Thirteenth International Conference on Learning
  Representations}.

\bibitem[{Xu et~al.(2024)Xu, Li, Xia, and Li}]{Enhancing2024xu}
Qiancheng Xu, Yongqi Li, Heming Xia, and Wenjie Li. 2024.
\newblock \href {https://doi.org/10.18653/v1/2024.findings-emnlp.561}
  {Enhancing tool retrieval with iterative feedback from large language
  models}.
\newblock In \emph{Findings of the Association for Computational Linguistics:
  EMNLP 2024}, pages 9609--9619, Miami, Florida, USA. Association for
  Computational Linguistics.

\bibitem[{Xu and Yan(2026)}]{Agent2026xu}
Renjun Xu and Yang Yan. 2026.
\newblock \href {https://openreview.net/forum?id=Er0p92BsmW} {Agent skills for
  large language models: Architecture, acquisition, security, and the path
  forward}.
\newblock In \emph{First Workshop on Agent Skills}.

\bibitem[{Yao et~al.(2023)Yao, Zhao, Yu, Du, Shafran, Narasimhan, and
  Cao}]{ReAct2022yao}
Shunyu Yao, Jeffrey Zhao, Dian Yu, Nan Du, Izhak Shafran, Karthik~R.
  Narasimhan, and Yuan Cao. 2023.
\newblock \href
  {http://dblp.uni-trier.de/db/conf/iclr/iclr2023.html#YaoZYDSN023} {React:
  Synergizing reasoning and acting in language models.}
\newblock In \emph{ICLR}. OpenReview.net.

\bibitem[{Zheng et~al.(2026)Zheng, Zhang, Ma, Yu, Zhu, Wu, Xu, Dong, Zhu,
  Huang, and Yu}]{SkillRouter2026zheng}
YanZhao Zheng, ZhenTao Zhang, Chao Ma, YuanQiang Yu, JiHuai Zhu, Yong Wu,
  Tianze Xu, Baohua Dong, Hangcheng Zhu, Ruohui Huang, and Gang Yu. 2026.
\newblock \href {https://arxiv.org/abs/2603.22455} {Skillrouter: Skill routing
  for llm agents at scale}.
\newblock \emph{Preprint}, arXiv:2603.22455.

\end{thebibliography}

\appendix

\section{Detailed Design of \textsc{SkillGrep}}
\label{app:SkillGrep}

\textsc{SkillGrep} is the retrieval tool used by \textsc{SkillBrowser}
to map an agent-generated retrieval intent to task-relevant skills.
This appendix describes its repository construction, query
representation, hybrid retrieval procedure, and compatibility-aware
selection mechanism.

\subsection{Retrieval-Oriented Skill Indexing}

\textsc{SkillGrep} operates over an indexed skill repository. Each skill
$s_j$ is represented as a pair
\begin{equation}
    s_j = (m_j, f_j),
\end{equation}
where $m_j$ is structured metadata used for retrieval and $f_j$ is the
full instructional content retained for downstream application. The
metadata includes a skill description, capability annotations, usage
conditions, representative positive examples, tags, interaction
information, and optional input/output information.

For each skill, we construct a collection of searchable views:
\begin{equation}
    \mathcal{V}_j =
    \{
    v_j^{\mathrm{desc}},
    v_j^{\mathrm{cap}},
    v_j^{\mathrm{use}},
    v_j^{\mathrm{ex}},
    v_j^{\mathrm{schema}},
    \{v_{j,l}^{\mathrm{sec}}\}_{l}
    \}.
\end{equation}
The description view contains the skill name, description, category,
and tags. The capability view contains explicit capability annotations.
The usage and example views represent when the skill should be used and
typical matching queries. The schema view, when available, represents
input/output characteristics. Finally, content-section views are
constructed from sections of the full instructional document.

Sections describing failure modes, contraindications, or conditions
under which a skill should not be used are not included as positive
content views. Instead, they are retained as negative evidence for
compatibility-aware scoring. This design preserves the complete skill
content while separating information that supports retrieval from
information that indicates non-applicability.

\subsection{Structured Retrieval Intent}

The input to \textsc{SkillGrep} is a structured retrieval intent generated
by the agent through the meta-skill. We denote it as
\begin{equation}
\begin{split}
    q_i = (&q_i^{\mathrm{text}}, q_i^{\mathrm{ctx}},
    C_i^{\mathrm{req}}, C_i^{\mathrm{des}}, I_i, O_i, \\
    &P_i^{+}, P_i^{-}, \Gamma_i),
\end{split}
\end{equation}
where $q_i^{\mathrm{text}}$ is a concise capability-oriented query,
$q_i^{\mathrm{ctx}}$ is task context, $C_i^{\mathrm{req}}$ and
$C_i^{\mathrm{des}}$ are required and desired capabilities, $I_i$ and
$O_i$ are input/output type hints, $P_i^{+}$ and $P_i^{-}$ are positive
and negative signals, and $\Gamma_i$ contains explicit constraints.

For lexical retrieval, all non-empty fields are concatenated into a
single query representation. For dense retrieval, the query is further
decomposed into semantic groups aligned with the skill views:
\begin{equation}
\begin{aligned}
    q_i^{\mathrm{desc}}
        &= [q_i^{\mathrm{text}}; q_i^{\mathrm{ctx}}], \\
    q_i^{\mathrm{cap}}
        &= [C_i^{\mathrm{req}}; C_i^{\mathrm{des}};
            q_i^{\mathrm{text}}], \\
    q_i^{\mathrm{use}}
        &= [P_i^{+}; q_i^{\mathrm{text}}], \\
    q_i^{\mathrm{schema}}
        &= [I_i; O_i; \Gamma_i], \\
    q_i^{\mathrm{sec}}
        &= [q_i^{\mathrm{ctx}}; q_i^{\mathrm{text}}].
\end{aligned}
\end{equation}
This alignment allows different parts of a retrieval intent to match the
corresponding information represented in each skill.

\subsection{Hybrid Candidate Retrieval}

\textsc{SkillGrep} combines three complementary retrieval signals.

\paragraph{Lexical Retrieval.}
We first concatenate the searchable views of each skill into a document
representation and compute a BM25 score:
\begin{equation}
    s_{\mathrm{lex}}(j)
    =
    \operatorname{BM25}
    \left(
        q_i,
        \operatorname{Concat}(\mathcal{V}_j)
    \right).
\end{equation}
This component captures exact matches such as method names, formula
names, software libraries, tools, and domain-specific terminology.

\paragraph{Sparse Multi-View Retrieval.}
We additionally compute TF--IDF cosine similarity separately for each
skill view:
\begin{equation}
    s_{\mathrm{sp}}^{v}(j)
    =
    \operatorname{Cos}
    \left(
        \operatorname{TFIDF}(q_i),
        \operatorname{TFIDF}(v_j)
    \right).
\end{equation}
The sparse view score is obtained by aggregating normalized per-view
scores:
\begin{equation}
    s_{\mathrm{sp}}(j)
    =
    \frac{
        \sum_{v \in \mathcal{V}_j}
        w_v \, \overline{s_{\mathrm{sp}}^{v}(j)}
    }{
        \sum_{v \in \mathcal{V}_j}
        w_v
    }.
\end{equation}
The current implementation assigns larger view weights to capability,
usage, example, and schema views than to content-section views:
\begin{equation}
    w_v =
    \begin{cases}
        1.4, & v = \mathrm{capability}, \\
        1.2, & v \in \{\mathrm{usage}, \mathrm{examples}\}, \\
        1.1, & v = \mathrm{schema}, \\
        0.7, & v = \mathrm{content\ section}, \\
        1.0, & \text{otherwise}.
    \end{cases}
\end{equation}

\paragraph{Dense Multi-View Retrieval.}
To retrieve semantically related skills whose wording differs from the
agent query, we embed query groups and corresponding skill views with a
\texttt{bge-base-en-v1.5} encoder. For each semantic group $g$, we compute
\begin{equation}
    s_{\mathrm{de}}^{g}(j)
    =
    \operatorname{Cos}
    \left(
        \operatorname{Enc}(q_i^{g}),
        \operatorname{Enc}(v_j^{g})
    \right).
\end{equation}
The aggregated dense score is
\begingroup
\small
\begin{equation}
\begin{split}
    s_{\mathrm{de}}(j) ={}&
    0.25s_{\mathrm{de}}^{\mathrm{desc}}(j)
    + 0.35s_{\mathrm{de}}^{\mathrm{cap}}(j)
    + 0.20s_{\mathrm{de}}^{\mathrm{use}}(j) \\
    &+ 0.10s_{\mathrm{de}}^{\mathrm{schema}}(j)
    + 0.10s_{\mathrm{de}}^{\mathrm{sec}}(j).
\end{split}
\end{equation}
\endgroup
In our implementation, dense retrieval uses a local Hugging Face encoder
with L2-normalized mean-pooled representations. The concrete model used
in experiments is reported in the experimental setup.

\subsection{Ranking Fusion and Candidate Recall}

Each retrieval path produces a ranked list of skills. We combine the
BM25 ranking, sparse-view rankings, and dense-view rankings using
reciprocal rank fusion:
\begin{equation}
    s_{\mathrm{rrf}}(j)
    =
    \sum_{\ell \in \mathcal{L}}
    \frac{1}{60 + \operatorname{rank}_{\ell}(j)},
\end{equation}
where $\mathcal{L}$ contains all available retrieval rankings.

The highest-ranked skills under $s_{\mathrm{rrf}}$ form an initial
candidate pool. In addition, skills that explicitly overlap with
requested capabilities or specified input/output types are added to the
pool. This design prevents structured requirements from being discarded
solely because they are not highly ranked by the general retrieval
signals.

\subsection{Compatibility-Aware Scoring}

For each recalled candidate, \textsc{SkillGrep} computes a final ranking
score that combines retrieval relevance with compatibility penalties:
\begingroup
\small
\begin{equation}
\begin{split}
    S(j) ={}&
    \alpha_{\mathrm{lex}} \bar{s}_{\mathrm{lex}}(j)
    + \alpha_{\mathrm{sp}} s_{\mathrm{sp}}(j)
    + \alpha_{\mathrm{de}} s_{\mathrm{de}}(j) \\
    &+ \alpha_{\mathrm{rrf}} \bar{s}_{\mathrm{rrf}}(j)
    + \alpha_{\mathrm{cap}} s_{\mathrm{cap}}(j)
    + \alpha_{\mathrm{use}} s_{\mathrm{use}}(j) \\
    &+ \alpha_{\mathrm{in}} s_{\mathrm{in}}(j)
    + \alpha_{\mathrm{out}} s_{\mathrm{out}}(j) \\
    &- \alpha_{\mathrm{pen}}
    \left(
        p_{\mathrm{contra}}(j)
        + p_{\mathrm{neg}}(j)
        + p_{\mathrm{hard}}(j)
    \right).
\end{split}
\end{equation}
\endgroup
Here, $\bar{s}_{\mathrm{lex}}$ and $\bar{s}_{\mathrm{rrf}}$ are normalized
lexical and fusion scores; $s_{\mathrm{cap}}$ and $s_{\mathrm{use}}$
capture explicit overlaps with capability annotations and usage
information; and $s_{\mathrm{in}}$ and $s_{\mathrm{out}}$ capture
input/output type overlap.

The three penalty terms have distinct roles. The contraindication
penalty $p_{\mathrm{contra}}$ is triggered when the query matches
conditions under which a skill should not be used. The negative-signal
penalty $p_{\mathrm{neg}}$ is triggered when a skill matches exclusions
expressed in the retrieval intent. The hard-mismatch penalty
$p_{\mathrm{hard}}$ accounts for explicit conflicts, including missing
required capabilities, incompatible input types, and violated
readability or execution constraints.

Under the default configuration in our implementation, the final score
uses
\begin{equation}
\begin{aligned}
    \alpha_{\mathrm{lex}} &= 0.2, &
    \alpha_{\mathrm{sp}} &= 0.1, &
    \alpha_{\mathrm{de}} &= 10.0, \\
    \alpha_{\mathrm{rrf}} &= 0.2, &
    \alpha_{\mathrm{pen}} &= 1.0,
\end{aligned}
\end{equation}
while the direct additive weights for capability, usage, and type
overlaps are set to zero:
\begin{equation}
    \alpha_{\mathrm{cap}}
    =
    \alpha_{\mathrm{use}}
    =
    \alpha_{\mathrm{in}}
    =
    \alpha_{\mathrm{out}}
    =
    0.
\end{equation}
These structured signals are nevertheless used for candidate inclusion,
compatibility diagnosis, and hard-mismatch detection.

\subsection{Selection and Abstention}

Candidates whose final scores do not exceed a configurable minimum
threshold are removed, and the remaining candidates are sorted by
$S(j)$. \textsc{SkillGrep} returns the top-$k$ ranked skill cards,
including their identifiers, descriptions, matched capability signals,
available sections, and score breakdowns. The full content associated
with a retrieved skill is incorporated in the subsequent skill-use
stage.

The tool abstains when no valid skill can satisfy an explicit hard
requirement. In particular, abstention is triggered when the repository
is empty, when all skills violate required input or interaction
constraints, when all returned candidates miss required capabilities, or
when no candidate remains above the score threshold. This mechanism
allows \textsc{SkillBrowser} to avoid injecting a retrieved skill when
the repository does not contain a suitable capability.

\section{Evaluation Benchmarks}
\label{app:benchmarks}

We evaluate on six subsets of SRA-Bench~\citep{SRA2026su}: \textsc{TheoremQA}, \textsc{LogicBench}, \textsc{CHAMP}, \textsc{MedCalcBench}, \textsc{BigCodeBench}, and \textsc{ToolQA}. 

SRA-Bench formulates each evaluation instance as a user query paired with one or 
more annotated gold skills. Each skill is a reusable capability artifact, including 
task-solving instructions, mathematical procedures, logical inference patterns, 
clinical calculation rules, or software-library usage guidance. Importantly, agents 
do not retrieve from a small task-specific library: the benchmark mixes gold skills 
with web-collected distractor skills, yielding a shared retrieval corpus of 26,262 
candidate skills. This setting evaluates whether an agent can identify the capability 
needed for solving a task under a realistically noisy skill library.

Table~\ref{tab:benchmark_statistics} summarizes the five original benchmark subsets, which contain 3,970 source instances and 622 source gold Skills. These source counts are not the valid-prediction denominators for every model or paired comparison. The additional ToolQA evaluation contains 100 valid trajectories per backbone (500 model--instance trajectories across five backbones); we do not add these repeated evaluations to the source-instance total. 
The selected subsets cover both knowledge-oriented skills, such as theorem application, 
logical inference, mathematical procedures, and medical calculators, and action-oriented 
skills, represented by software-library usage in \textsc{BigCodeBench}.

\begin{table*}[t]
\centering
\small
\setlength{\tabcolsep}{5pt}
\renewcommand{\arraystretch}{1.15}
\begin{tabular}{l l r r l}
\toprule
\textbf{Benchmark} & \textbf{Capability Type} & \textbf{\# Inst.} & 
\textbf{\# Skills} & \textbf{Evaluation} \\
\midrule
TheoremQA     & Theorem application         & 747   & 320 & Single, rule-based \\
LogicBench    & Logical reasoning patterns  & 760   & 19  & Single, rule-based \\
CHAMP         & Mathematical concepts       & 223   & 89  & Multi, rule-based \\
MedCalcBench & Medical calculators         & 1,100 & 55  & Single, rule-based \\
BigCodeBench  & Software libraries          & 1,140 & 139 & Multi, execution-based \\
\midrule
\textbf{Subtotal} & -- & \textbf{3,970} & \textbf{622} & -- \\
\bottomrule
\end{tabular}
\caption{Source statistics for the five original SRA-Bench subsets; ToolQA evaluation counts are reported separately. 
The complete retrieval corpus additionally contains web-collected distractor skills, 
resulting in 26,262 candidate skills in total.}
\label{tab:benchmark_statistics}
\end{table*}

\subsection{TheoremQA}

\textsc{TheoremQA} evaluates the retrieval and application of reusable mathematical 
and scientific principles. Its questions span domains such as combinatorics, calculus, 
linear algebra, physics, probability, economics, and graph theory. Each instance is 
paired with a single gold skill describing the theorem, formula, or canonical procedure 
needed to solve the question. For example, an instance asking for the number of ways 
to divide a set of eight elements into five non-empty ordered subsets is paired with 
the skill \textit{Lah Numbers}. This subset therefore measures whether the agent can 
retrieve an appropriate abstract principle rather than relying only on surface-level 
lexical overlap.

\subsection{LogicBench}

\textsc{LogicBench} evaluates retrieval for deductive reasoning tasks. Each instance 
contains a collection of premises and several candidate conclusions, and the model 
must identify which conclusion logically follows from the premises. The associated 
skills represent reusable logical reasoning patterns rather than domain-specific facts. 
Since the benchmark uses single-skill annotations and rule-based answer checking, it 
provides a controlled setting for testing whether retrieved reasoning procedures can 
support symbolic inference.

\subsection{CHAMP}

\textsc{CHAMP} contains mathematical problem-solving instances that require applying 
one or more mathematical concepts or procedures. Compared with \textsc{TheoremQA}, 
which is annotated with a single theorem-oriented skill, \textsc{CHAMP} is configured 
as a multi-skill benchmark in SRA-Bench. Its problems include combinatorial counting 
and other structured mathematical reasoning tasks. Consequently, this subset evaluates 
whether a retrieval method can recover multiple complementary skills when solving a 
single query requires composing several procedures.

\subsection{MedCalcBench}

\textsc{MedCalcBench} evaluates retrieval of clinical calculation procedures. Each 
query describes patient attributes such as age, sex, laboratory measurements, or body 
weight, and asks for the result of a named or implied medical calculator. For example, 
a released instance requests estimated creatinine clearance for a 68-year-old female 
patient using the Cockcroft--Gault equation. The benchmark is especially relevant for 
skill retrieval because correct task execution depends on identifying the appropriate 
calculator, grounding its required variables, and applying the formula consistently. 
We use this benchmark only as a procedural calculation task and do not interpret model 
outputs as medical advice.

\subsection{BigCodeBench}

\textsc{BigCodeBench} evaluates retrieval of software-library usage skills. Unlike the 
other selected subsets, its evaluation is execution-based: a generated solution is 
checked by running the associated tests rather than by comparing a normalized textual 
answer. SRA-Bench models this subset as multi-skill retrieval, reflecting the fact that 
a programming task may require combining several API or library capabilities. This 
benchmark therefore tests whether retrieved skills can support executable task 
completion, complementing the rule-based reasoning benchmarks above.

\subsection{ToolQA}
ToolQA adds multi-step tool interaction to the evaluation. A capability-oriented query can identify an intermediate tool protocol not well represented by the initial user request. We evaluate 100 valid trajectories per model and report accuracy, gold-Skill recall, search frequency, LLM calls, and resource use in Appendices~\ref{app:toolqa} and~\ref{app:efficiency}.

\subsection{Example Instances}

Table~\ref{tab:benchmark_examples_full_query} presents representative instances from the five
SRA-Bench subsets used in our evaluation. The examples illustrate the diversity of
capabilities covered by the benchmark: mathematical theorem application, logical
reasoning, combinatorial problem solving, clinical calculation, and software-library
usage. While \textsc{TheoremQA}, \textsc{LogicBench}, and \textsc{MedCalcBench}
associate each query with a single target skill, \textsc{BigCodeBench} may require
multiple complementary skills, as illustrated by its example involving both
\texttt{random} and \texttt{itertools}.

\onecolumn
\begin{footnotesize}
\begin{longtable}{@{}p{0.12\textwidth} p{0.64\textwidth} p{0.18\textwidth}@{}}
\label{tab:benchmark_examples_full_query} \\
\toprule
\textbf{Benchmark} & \textbf{Query Example} & \textbf{Gold Skill Name} \\
\midrule
\endfirsthead

\multicolumn{3}{l}{\small\itshape Table~\thetable{} continued from the previous page.} \\
\toprule
\textbf{Benchmark} & \textbf{Query Example} & \textbf{Gold Skill Name} \\
\midrule
\endhead

\midrule
\multicolumn{3}{r}{\small\itshape Continued on the next page.} \\
\endfoot

\bottomrule
\endlastfoot

TheoremQA &
How many ways are there to divide a set of 8 elements into 5 non-empty ordered subsets? &
Lah Numbers \\
\addlinespace[5pt]
LogicBench &
If an individual decides to take a temporary break from their professional responsibilities, it can contribute to their overall sense of rejuvenation. however, if they choose to burden themselves with additional tasks, they may find themselves feeling overwhelmed. in this scenario, we are aware that at least one of the following statements holds true: (1) mike takes a break from work, or (2) he do not experience overwhelming feelings. it is important to note that we lack knowledge regarding which one, or both, of these statements are accurate.

In light of the context provided, what conclusion can be considered the most appropriate?
  choice\_1: Luke takes a break from work and feels refreshed.
  choice\_2: Mike takes on more tasks and feels overwhelmed.
  choice\_3: Mike takes a break from work and feels overwhelmed.
  choice\_4: he will feel refreshed or he doesn't take on more tasks &
Bidirectional Dilemma \\
\addlinespace[5pt]
CHAMP &
Let a string consist of digit 1, 2, 3. How many such strings of length 6 have adjacent digit differing by less than or equal to 1? &
Rule of Sum (Addition Principle) \\

\addlinespace[5pt]
BigCodeBench &
Calculates the average of the sums of absolute differences between each pair of consecutive numbers for all permutations of a given list. Each permutation is shuffled before calculating the differences. Args: -- \texttt{numbers} (\texttt{list}): A list of numbers. Default is numbers from 1 to 10.
\newline
The function should output with:
\newline
\hspace*{1em}\texttt{float}: The average of the sums of absolute differences for each shuffled permutation of the list.
\newline
You should write self-contained code starting with:
\newline
\texttt{import itertools}
\newline
\texttt{from random import shuffle}
\newline
\texttt{def task\_func(numbers=list(range(1, 3))):}
&
\texttt{random} -- Core API and Common Patterns; \newline
\texttt{itertools} -- Iterator Building Blocks \\

\addlinespace[5pt]
MedCalcBench &
Patient Note:
An 87-year-old man was admitted to our hospital for anorexia for several days, high-grade fever from the previous day, and liver dysfunction. Of note, he had a history of hypertension, diabetes mellitus (DM), and angina. Physical examination findings included: clear consciousness; height, 163 cm; weight, 48 kg; blood pressure, 66/40 mmHg; heart rate, 75/min; respiratory rate, 22/min; oxygen saturation of peripheral artery, 96\%; and body temperature, 38.1 °C. He had no surface lymphadenopathy. Laboratory findings included: white blood cell (WBC) count, 4.2 × 109/L; hemoglobin, 9.6 g/dL; platelet count, 106 × 109/L; lactate dehydrogenase (LDH), 1662 IU/L; aspartate aminotransferase (AST), 6562 IU/L; alanine aminotransferase (ALT), 1407 IU/L; alkaline phosphatase (ALP), 509 IU/L; $\gamma$-glutamyl transpeptidase ($\gamma$-GTP), 130 IU/L; total bilirubin, 2.7 mg/dL; prothrombin time (PT), 20.3 s; international normalized ratio (INR), 1.73; blood urea nitrogen (BUN), 34.4 mg/dL; and creatinine, 1.4 mg/dL. Serologic tests were all negative for hepatitis A, B, C, and E. Computed tomography (CT) performed at admission revealed no distinct nodules or masses in the liver, but mild splenomegaly, multiple hypodense splenic lesions, and para-aortic lymph node swelling (Fig. a–c). Antibiotics were administered intravenously for suspected acute cholangitis and sepsis, but the fever persisted. Abdominal ultrasonography (US) and magnetic resonance imaging (MRI) performed 2 days later newly depicted gallbladder wall thickening and ascites (Figs. , ). On the same night, the patient presented with grade II hepatic encephalopathy, with blood tests showing a decrease in platelet count and worsening liver function suggestive of “acute type” ALF with hepatic coma and disseminated intravascular coagulation (DIC). While consideration was being given to treating acute liver failure, the patient suddenly developed ventricular fibrillation and died despite immediate resuscitation measures on day 3 of hospitalization.
Given the rapid clinical course, an autopsy was performed in this patient with suspected ALF likely due to drug-induced liver injury, autoimmune hepatitis, hepatitis of indeterminate etiology, circulatory disturbance, metabolic disease, or malignant cells including malignant lymphoma. This led to the patient being diagnosed with malignant, diffuse large B-cell lymphoma, characterized by multiple lymph-node, as well as multi-organ (i.e., spleen, liver, lung, and prostate) involvement: (1) general swelling of lymph nodes in the para-abdominal aorta (Fig. a), peripancreatic, peribronchial, pulmonary hilar, subclavicular regions, as well as in the pelvis, especially around the iliac arteries, measuring up to 60 mm in diameter; (2) presence of multiple whitish splenic nodules (Fig. b) with the entire spleen shown to be microscopically infiltrated by tumor cells (Fig. a); (3) irregular geographical morphology of the liver (Fig. c) with the lymphoma cells mainly infiltrating the portal areas (Fig. b); (4) involvement of the right lobe of the lung; (5) involvement of both lobes of the prostate consequent to outer connective tissue involvement (Fig. c). Microscopically, the tumor cells in the lymph nodes were shown to have medium-sized irregular nuclei and small nucleoli with a high nuclear cytoplasmic ratio, thus showing diffuse proliferation characterized by the presence of large numbers of small T-cells (Fig. ). Immunohistochemistry revealed the tumor cells to be positive for CD20 (Fig. a), CD79a (Fig. b), bcl-2 (Fig. c), and MUM1 (Fig. d), but negative for CD3 (Fig. e), CD10, bcl-6, and cyclin D1, with a Ki-67 labeling index (Ki-67 LI) of 16\% (Fig. f). The causes of death were thus determined as (1) DIC and (2) ALF associated with diffuse large B-cell lymphoma.

Task:
What is the patient's Creatinine Clearance using the Cockroft-Gault Equation in terms of mL/min? You should use the patient's adjusted body weight in kg instead of the patient's actual body weight if the patient is overweight or obese based on their BMI. If the patient's BMI's normal, set their adjusted body weight to the minimum of the ideal body and actual weight. If the patient is underweight, please set their adjusted body weight to their actual body weight. You should use the patient's medical values and health status when they were first admitted to the hospital prior to any treatment. &
Creatinine Clearance (Cockcroft--Gault Equation) \\

\bottomrule
\caption{Representative instances and their annotated gold skills from the five
SRA-Bench subsets used in our evaluation. Complete queries are retained without
truncation.}
\end{longtable}
\end{footnotesize}

\twocolumn

\section{Full Results of Selective Skill Exposure Analysis}
\label{app:selective_exposure}

We provide the complete model--benchmark level results for the Selective Skill Exposure analysis. For each model--benchmark pair, we construct two subsets according to the outcomes under the no-Skill and gold-Skill reference conditions. An instance is \emph{skill-beneficial} if the model answers incorrectly without external Skills but correctly when provided with the annotated gold Skill. An instance is \emph{skill-unnecessary} if the model answers correctly under both conditions. This partition is constructed separately for each model--benchmark pair because the utility of external procedural support depends on the backbone model's native capability.

We compare \textbf{Meta-Skill}, which denotes Meta-Skill-managed Skill access, with \textbf{SkillGrep}, which statically retrieves Skills through the same hybrid retrieval backend. On skill-beneficial instances, we report Beneficial Exposure Rate (BER), i.e., the percentage of instances on which retrieved Skill contents are exposed, together with downstream accuracy. On skill-unnecessary instances, we report Unnecessary Exposure Rate (UER), where lower values indicate less unnecessary procedural exposure.

\begin{table*}[t]
\centering
\footnotesize
\setlength{\tabcolsep}{5.0pt}
\renewcommand{\arraystretch}{1.08}
\begin{tabular}{@{}llrrrrr@{}}
\toprule
\textbf{Model} & \textbf{Benchmark} & \textbf{\#Ben.}
& \multicolumn{2}{c}{\textbf{BER} $\uparrow$}
& \multicolumn{2}{c}{\textbf{Accuracy} $\uparrow$} \\
\cmidrule(lr){4-5}\cmidrule(l){6-7}
& & & \textbf{Meta-Skill} & \textbf{SkillGrep}
& \textbf{Meta-Skill} & \textbf{SkillGrep} \\
\midrule

\multirow{4}{*}{Qwen3-8B}
& \textsc{CHAMP}        & 59  & 59.3  & 100.0 & 67.8 & \textbf{78.0} \\
& \textsc{LogicBench}   & 144 & 77.8  & 100.0 & \textbf{53.5} & 50.0 \\
& \textsc{MedCalcBench} & 444 & 64.0  & 100.0 & 60.4 & \textbf{64.2} \\
& \textsc{TheoremQA}    & 212 & 100.0 & 100.0 & \textbf{90.1} & 81.6 \\
\midrule

\multirow{4}{*}{Qwen3.6-Plus}
& \textsc{CHAMP}        & 15  & 26.7  & 100.0 & 33.3 & \textbf{53.3} \\
& \textsc{LogicBench}   & 123 & 59.3  & 100.0 & 52.8 & \textbf{64.2} \\
& \textsc{MedCalcBench} & 150 & 97.3  & 100.0 & \textbf{94.7} & 88.7 \\
& \textsc{TheoremQA}    & 48  & 97.9  & 100.0 & \textbf{91.7} & 83.3 \\
\midrule

\multirow{4}{*}{GLM-5.1}
& \textsc{CHAMP}        & 32  & 43.8  & 100.0 & 65.6 & \textbf{78.1} \\
& \textsc{LogicBench}   & 195 & 63.6  & 100.0 & \textbf{70.8} & 69.7 \\
& \textsc{MedCalcBench} & 245 & 100.0 & 100.0 & \textbf{97.6} & 82.4 \\
& \textsc{TheoremQA}    & 84  & 100.0 & 100.0 & \textbf{90.5} & 78.6 \\
\midrule

\multirow{4}{*}{\makecell[l]{GPT-5.4-\\mini}}
& \textsc{CHAMP}        & 24  & 8.3  & 100.0 & 62.5 & \textbf{70.8} \\
& \textsc{LogicBench}   & 70  & 4.3  & 100.0 & 11.4 & \textbf{64.3} \\
& \textsc{MedCalcBench} & 22  & 54.5 & 100.0 & \textbf{81.8} & 77.3 \\
& \textsc{TheoremQA}    & 29  & 93.1 & 100.0 & 89.7 & 89.7 \\
\midrule

\multirow{4}{*}{\makecell[l]{Gemini-3.1-\\Flash-Lite}}
& \textsc{CHAMP}        & 12  & 0.0   & 100.0 & 0.0  & \textbf{83.3} \\
& \textsc{LogicBench}   & 71  & 4.2   & 100.0 & 2.8  & \textbf{66.2} \\
& \textsc{MedCalcBench} & 37  & 100.0 & 100.0 & 81.1 & 81.1 \\
& \textsc{TheoremQA}    & 21  & 38.1  & 100.0 & 23.8 & \textbf{85.7} \\

\bottomrule
\end{tabular}
\caption{
Full results on \emph{skill-beneficial} instances. These instances are answered incorrectly without external Skills but correctly when the annotated gold Skill is provided. \textbf{BER} denotes Beneficial Exposure Rate, i.e., the percentage of instances on which retrieved Skill contents are exposed. 
}
\label{tab:full_beneficial_exposure}
\end{table*}

\begin{table*}[t]
\centering
\footnotesize
\setlength{\tabcolsep}{5.0pt}
\renewcommand{\arraystretch}{1.08}
\begin{tabular}{@{}llrrrrr@{}}
\toprule
\textbf{Model} & \textbf{Benchmark} & \textbf{\#Unn.}
& \multicolumn{2}{c}{\textbf{UER} $\downarrow$}
& \multicolumn{2}{c}{\textbf{Accuracy} $\uparrow$} \\
\cmidrule(lr){4-5}\cmidrule(l){6-7}
& & & \textbf{Meta-Skill} & \textbf{SkillGrep}
& \textbf{Meta-Skill} & \textbf{SkillGrep} \\
\midrule

\multirow{4}{*}{Qwen3-8B}
& \textsc{CHAMP}        & 90  & 36.7 & 100.0 & \textbf{94.4} & 91.1 \\
& \textsc{LogicBench}   & 517 & 72.9 & 100.0 & \textbf{92.8} & 83.0 \\
& \textsc{MedCalcBench} & 381 & 45.4 & 100.0 & \textbf{97.6} & 86.9 \\
& \textsc{TheoremQA}    & 228 & 99.1 & 100.0 & \textbf{96.9} & 86.0 \\
\midrule

\multirow{4}{*}{Qwen3.6-Plus}
& \textsc{CHAMP}        & 173 & 31.2 & 100.0 & \textbf{98.8} & 97.7 \\
& \textsc{LogicBench}   & 565 & 52.0 & 100.0 & \textbf{98.6} & 89.4 \\
& \textsc{MedCalcBench} & 859 & 90.6 & 100.0 & \textbf{99.3} & 97.8 \\
& \textsc{TheoremQA}    & 576 & 96.5 & 100.0 & \textbf{99.3} & 98.6 \\
\midrule

\multirow{4}{*}{GLM-5.1}
& \textsc{CHAMP}        & 147 & 61.2 & 100.0 & 91.2 & \textbf{94.6} \\
& \textsc{LogicBench}   & 487 & 68.4 & 100.0 & \textbf{97.5} & 91.2 \\
& \textsc{MedCalcBench} & 714 & 99.4 & 100.0 & \textbf{98.9} & 95.8 \\
& \textsc{TheoremQA}    & 504 & 99.4 & 100.0 & \textbf{100.0} & 98.0 \\
\midrule

\multirow{4}{*}{\makecell[l]{GPT-5.4-\\mini}}
& \textsc{CHAMP}        & 146 & 13.7 & 100.0 & \textbf{99.3} & 95.9 \\
& \textsc{LogicBench}   & 186 & 5.9  & 100.0 & \textbf{97.8} & 88.7 \\
& \textsc{MedCalcBench} & 226 & 56.6 & 100.0 & \textbf{99.1} & 94.2 \\
& \textsc{TheoremQA}    & 205 & 90.2 & 100.0 & \textbf{99.0} & 95.6 \\
\midrule

\multirow{4}{*}{\makecell[l]{Gemini-3.1-\\Flash-Lite}}
& \textsc{CHAMP}        & 147 & 10.2 & 100.0 & \textbf{97.3} & 91.8 \\
& \textsc{LogicBench}   & 192 & 7.3  & 100.0 & \textbf{97.9} & 89.6 \\
& \textsc{MedCalcBench} & 205 & 100.0 & 100.0 & 94.1 & \textbf{96.1} \\
& \textsc{TheoremQA}    & 228 & 44.3 & 100.0 & \textbf{100.0} & 96.1 \\

\bottomrule
\end{tabular}
\caption{
Full results on \emph{skill-unnecessary} instances. These instances are answered correctly both without external Skills and with the annotated gold Skill. \textbf{UER} denotes Unnecessary Exposure Rate, i.e., the percentage of instances on which retrieved Skill contents are exposed although they are not necessary for correctness under the evaluated model. 
}
\label{tab:full_unnecessary_exposure}
\end{table*}

\paragraph{Analysis.}
Tables~\ref{tab:full_beneficial_exposure} and~\ref{tab:full_unnecessary_exposure} provide a fine-grained view of how Meta-Skill-managed access affects Skill exposure. On skill-unnecessary instances, the advantage is highly consistent: Meta-Skill-managed access lowers unnecessary exposure relative to static SkillGrep retrieval in 19 of 20 model--benchmark pairs (with one tie), and improves downstream accuracy in 18 out of 20 pairs. Micro-aggregated across all settings, it reduces UER from $100.0\%$ to $70.9\%$ while improving accuracy from $93.2\%$ to $97.9\%$. These results support the central motivation of Selective Skill Exposure: blindly injecting retrieved procedural contents can disrupt tasks already solvable without external assistance.

On skill-beneficial instances, Meta-Skill-managed access exhibits a more task- and model-dependent pattern. It achieves strong gains on several \textsc{MedCalcBench} and \textsc{TheoremQA} settings, including improvements of $15.2$ points on GLM-5.1/\textsc{MedCalcBench} and $11.9$ points on GLM-5.1/\textsc{TheoremQA}. These results indicate that capability-oriented access can improve the utility of exposed Skills when the agent successfully articulates its need. However, substantial drops on GPT-5.4-mini and Gemini-3.1-Flash-Lite for \textsc{LogicBench} and \textsc{CHAMP} show that some agents do not reliably invoke Skill access when external support would be beneficial. Overall, Meta-Skill management provides a robust mechanism for avoiding unnecessary Skill exposure, while improving the recognition of beneficial Skill use remains an important direction for future work.

\clearpage
\section{Static Retrieval and Retriever Ablations}
\label{app:retrieval_evaluation}

\paragraph{Full-corpus static comparison.}
Table~\ref{tab:static_retrievers} compares retrieval from the dataset task description over the same 26,262-Skill repository. We report GoldHit@$5$, GoldMRR@$5$, Recall@$1/5$, and nDCG@$1/5$ using the benchmark evaluation. SkillGrep leads all reported metrics, supporting its use as a strong static baseline for the exposure comparison.

\begin{table*}[t]
\centering
\footnotesize
\setlength{\tabcolsep}{4pt}
\renewcommand{\arraystretch}{1.12}
\begin{tabular}{@{}lrrrrrr@{}}
\toprule
Retriever & GoldHit@5 & GoldMRR@5 & Recall@1 & Recall@5 & nDCG@1 & nDCG@5 \\
\midrule
SKILLGREP & \textbf{0.62} & \textbf{0.50} & \textbf{0.40} & \textbf{0.59} & \textbf{0.43} & \textbf{0.51} \\
SR-Agent BGE Retriever & 0.47 & 0.37 & 0.31 & 0.45 & 0.31 & 0.39 \\
SR-Agent TF-IDF & 0.39 & 0.29 & 0.22 & 0.37 & 0.23 & 0.30 \\
SR-Agent BM25 & 0.49 & 0.36 & 0.28 & 0.46 & 0.30 & 0.38 \\
SR-Agent Contriever & 0.41 & 0.31 & 0.24 & 0.39 & 0.24 & 0.32 \\
\bottomrule
\end{tabular}
\caption{Static retrieval over the full Skill corpus using original task descriptions as queries. Values are proportions, reported at the reported precision; bold marks the best result.}
\label{tab:static_retrievers}
\end{table*}

\paragraph{Leave-one-component-out ablation.}
Table~\ref{tab:retriever_ablation} removes individual SkillGrep components for offline gold-Skill retrieval at the same $k=5$. Keeping the ranking cutoff fixed avoids comparing thresholded single-channel scores that require separate calibration. Removing dense retrieval or capability/type signals causes the largest GoldHit@$5$ drops; removing sparse views and the RRF score also reduces retrieval quality. The penalty has little measurable effect. This experiment evaluates retrieval, not downstream end-task accuracy.

Capability/type signals participate in candidate inclusion and compatibility checks even though their direct additive score weights are zero in the default configuration (Appendix~\ref{app:SkillGrep}). Thus, their ablation is not equivalent to setting those already-zero coefficients to zero. The ``No RRF Score'' variant is reported separately from the broader use of rankings for candidate recall.

\begin{table*}[t]
\centering
\footnotesize
\setlength{\tabcolsep}{4pt}
\renewcommand{\arraystretch}{1.12}
\begin{tabular}{@{}lrrrrrr@{}}
\toprule
Variant & GoldHit@5 & GoldMRR@5 & Recall@1 & Recall@5 & nDCG@1 & nDCG@5 \\
\midrule
Full SKILLGREP & \textbf{0.623} & \textbf{0.500} & \textbf{0.402} & \textbf{0.590} & \textbf{0.429} & \textbf{0.511} \\
No Penalty & \textbf{0.623} & \textbf{0.500} & 0.401 & \textbf{0.590} & \textbf{0.429} & \textbf{0.511} \\
No RRF Score & 0.578 & 0.471 & 0.381 & 0.544 & 0.408 & 0.477 \\
No Sparse View & 0.548 & 0.446 & 0.359 & 0.514 & 0.383 & 0.452 \\
No Dense & 0.463 & 0.422 & 0.364 & 0.436 & 0.392 & 0.413 \\
No Capability/Type Signals & 0.462 & 0.431 & 0.379 & 0.436 & 0.407 & 0.420 \\
\bottomrule
\end{tabular}
\caption{Leave-one-component-out SkillGrep ablation. All values are proportions; bold marks the best results, including ties at the reported precision.}
\label{tab:retriever_ablation}
\end{table*}

\section{ToolQA Retrieval and Interaction}
\label{app:toolqa}

Table~\ref{tab:toolqa_access} reports ToolQA retrieval behavior. Search is the fraction of trajectories invoking retrieval; Recall@$5$ is evaluated across all trajectories, and R@$5$/Search is the conditional rate among searching trajectories. Static methods always retrieve but obtain zero gold-Skill recall. SkillBrowser's capability-oriented query yields conditional recall between $0.54$ and $1.00$ in the reported runs, with one retrieval call per searching trajectory. This supports alignment with intermediate tool-use needs without demonstrating iterative retrieval refinement.

For GLM-5.1, conditional recall is $1.00$, but only $34\%$ of trajectories search; high retrieval quality when invoked therefore coexists with missed access opportunities. All models remain below Gold Exposure in ToolQA accuracy. That gap is consistent with under-invocation, retrieval misses, and downstream execution failures, but these aggregate metrics do not identify a causal breakdown of the errors.

\begin{table*}[t]
\centering
\footnotesize
\setlength{\tabcolsep}{4pt}
\renewcommand{\arraystretch}{1.12}
\begin{tabular}{@{}llrrrr@{}}
\toprule
Model & Method & Accuracy & Recall@5 & Search & R@5/Search \\
\midrule
Qwen3-8B & Static BM25 & 0.17 & 0.00 & 1.00 & 0.00 \\
 & Static SKILLGREP & 0.17 & 0.00 & 1.00 & 0.00 \\
 & SKILLBROWSER & 0.44 & 0.50 & 0.93 & 0.54 \\
 & Gold Exposure & 0.51 & - & - & - \\
Qwen3.6-Plus & Static BM25 & 0.32 & 0.00 & 1.00 & 0.00 \\
 & Static SKILLGREP & 0.27 & 0.00 & 1.00 & 0.00 \\
 & SKILLBROWSER & 0.57 & 0.66 & 0.90 & 0.73 \\
 & Gold Exposure & 0.76 & - & - & - \\
GLM-5.1 & Static BM25 & 0.37 & 0.00 & 1.00 & 0.00 \\
 & Static SKILLGREP & 0.39 & 0.00 & 1.00 & 0.00 \\
 & SKILLBROWSER & 0.51 & 0.34 & 0.34 & 1.00 \\
 & Gold Exposure & 0.73 & - & - & - \\
GPT-5.4-mini & Static BM25 & 0.33 & 0.00 & 1.00 & 0.00 \\
 & Static SKILLGREP & 0.32 & 0.00 & 1.00 & 0.00 \\
 & SKILLBROWSER & 0.43 & 0.40 & 0.66 & 0.61 \\
 & Gold Exposure & 0.71 & - & - & - \\
Gemini-3.1-Flash-Lite & Static BM25 & 0.17 & 0.00 & 1.00 & 0.00 \\
 & Static SKILLGREP & 0.18 & 0.00 & 1.00 & 0.00 \\
 & SKILLBROWSER & 0.61 & 0.88 & 1.00 & 0.88 \\
 & Gold Exposure & 0.70 & - & - & - \\
\bottomrule
\end{tabular}
\caption{ToolQA accuracy and retrieval behavior (proportions), with 100 valid trajectories per model. Dashes indicate that search metrics do not apply to direct gold exposure.}
\label{tab:toolqa_access}
\end{table*}

\section{Resource Use and Latency}
\label{app:efficiency}

\paragraph{Retrieval calls.}
The five ToolQA runs contain 500 valid model--instance trajectories in total. They contain 383 SkillGrep calls ($76.6\%$ search rate, $0.766$ calls per trajectory). Each searching trajectory makes one call and loads five candidate Skills, averaging $3.83$ loaded Skills per trajectory. Of these calls, 278 retrieve a gold Skill ($72.6\%$). The conditional aggregate is computed from counts, rather than averaging the five model-specific conditional rates.

\paragraph{LLM calls.}
Table~\ref{tab:toolqa_llm_calls} counts all LLM calls along the tool-interaction trajectory. SkillBrowser requires more calls than LLM Selection for four backbones and fewer for GLM-5.1. Its cost depends on both retrieval invocation and subsequent tool use; one SkillGrep call does not imply only one additional LLM call.

\begin{table*}[t]
\centering
\footnotesize
\setlength{\tabcolsep}{4pt}
\renewcommand{\arraystretch}{1.12}
\begin{tabular}{@{}lrrr@{}}
\toprule
Model & SkillBrowser & Gold Exposure & LLM Selection \\
\midrule
Qwen3-8B & 8.66 & 4.30 & 5.30 \\
Qwen3.6-Plus & 8.60 & 6.80 & 6.90 \\
GLM-5.1 & 5.35 & 3.30 & 6.70 \\
GPT-5.4-mini & 5.96 & 3.30 & 5.50 \\
Gemini-3.1-Flash-Lite & 5.40 & 2.20 & 4.80 \\
\bottomrule
\end{tabular}
\caption{Average total LLM calls per ToolQA instance.}
\label{tab:toolqa_llm_calls}
\end{table*}

\paragraph{Tokens and latency.}
Tables~\ref{tab:glm_tokens} and~\ref{tab:glm_latency} report average complete-trajectory input/output token counts and latency for GLM-5.1. Token counts include the trajectory's repeated model interactions rather than only the final answer. Combined token use is lower than LLM Selection on TheoremQA, MedCalcBench, and BigCodeBench, and higher on LogicBench, CHAMP, and ToolQA. Latency is higher on all six benchmarks. SkillBrowser's online path includes query generation, retrieval, and Skill loading; Gold Exposure directly loads the annotated Skill, whereas LLM Selection starts from a precomputed BM25 top-50 set and excludes BM25 retrieval from its reported latency. These measurements characterize the evaluated configurations rather than equal-stage online latency. Input/output tokens provide a pricing-independent resource proxy, but not a universal monetary-cost estimate.

\begin{table*}[t]
\centering
\footnotesize
\setlength{\tabcolsep}{4pt}
\renewcommand{\arraystretch}{1.12}
\begin{tabular}{@{}lrrr@{}}
\toprule
Benchmark & SkillBrowser & Gold Exposure & LLM Selection \\
\midrule
TheoremQA & 3014.1 / 947.7 & 1095.6 / 419.4 & 4035.1 / 411.9 \\
LogicBench & 6014.4 / 1714.3 & 1392.1 / 527.3 & 4661.2 / 563.5 \\
CHAMP & 3372.4 / 3850.1 & 1857.9 / 693.7 & 6319.6 / 879.3 \\
MedCalcBench & 3395.4 / 797.6 & 1489.8 / 278.6 & 6120.9 / 293.6 \\
ToolQA & 44569.1 / 798.5 & 16279.7 / 214.1 & 40386.3 / 982.5 \\
BigCodeBench & 1627.2 / 627.7 & 4827.5 / 272.4 & 4492.0 / 301.9 \\
\bottomrule
\end{tabular}
\caption{Average GLM-5.1 input / output tokens per instance over the complete trajectory.}
\label{tab:glm_tokens}
\end{table*}
\begin{table*}[t]
\centering
\footnotesize
\setlength{\tabcolsep}{4pt}
\renewcommand{\arraystretch}{1.12}
\begin{tabular}{@{}lrrr@{}}
\toprule
Benchmark & SkillBrowser & Gold Exposure & LLM Selection \\
\midrule
TheoremQA & 23.384 & 8.601 & 10.055 \\
LogicBench & 42.356 & 14.203 & 14.582 \\
CHAMP & 63.435 & 10.357 & 16.785 \\
MedCalcBench & 31.063 & 9.881 & 11.148 \\
ToolQA & 36.937 & 18.500 & 33.139 \\
BigCodeBench & 19.050 & 5.649 & 7.427 \\
\bottomrule
\end{tabular}
\caption{Average GLM-5.1 latency in seconds per instance. LLM Selection excludes precomputed BM25 retrieval; SkillBrowser includes online retrieval and Skill loading.}
\label{tab:glm_latency}
\end{table*}

\section{Correctness Transitions Relative to Gold Exposure}
\label{app:gold_transitions}

To examine why SkillBrowser can exceed Gold Exposure in some settings, we compare per-instance correctness across six datasets and five models. Table~\ref{tab:gold_transitions} reports macro-averages across the 30 settings. The four transition proportions sum to $99.9\%$ because of rounding. These equal-setting macro-averages differ from the valid-pair pooled statistics in Appendix~\ref{app:statistics}; in particular, $5.0\%-9.9\%$ is not the pooled accuracy difference.

Among Gold-incorrect/SkillBrowser-correct instances, the reported search and gold-exposure rates are $60.1\%$ and $51.6\%$, respectively. For the reverse transition they are $22.9\%$ and $14.8\%$. Improvements therefore often involve retrieval, rather than simply avoiding Skills, while losses more often co-occur with failure to search or expose a gold Skill. These are descriptive associations: they do not establish whether any particular search, retrieved Skill, or skipped exposure caused the correctness transition.

\begin{table*}[t]
\centering
\footnotesize
\setlength{\tabcolsep}{4pt}
\renewcommand{\arraystretch}{1.12}
\begin{tabular}{@{}llr@{}}
\toprule
Gold Exposure & SkillBrowser & Share (\%) \\
\midrule
Incorrect & Correct & 5.0 \\
Correct & Correct & 72.1 \\
Correct & Incorrect & 9.9 \\
Incorrect & Incorrect & 12.9 \\
\bottomrule
\end{tabular}
\caption{Correctness transitions, macro-averaged across 30 model--dataset settings.}
\label{tab:gold_transitions}
\end{table*}

\section{Paired Statistical Comparisons}
\label{app:statistics}

For each comparator, paired analysis retains the intersection of valid predictions. Let $b$ count instances correct only for SkillBrowser and $c$ count instances correct only for the comparator. A two-sided exact McNemar test uses the binomial null $b\mid(b+c)\sim\operatorname{Binomial}(b+c,1/2)$. The discordant counts yield the $p$-values in Table~\ref{tab:paired_statistics}. Aggregate results favor SkillBrowser over both deployable baselines but favor Gold Exposure over SkillBrowser. Different valid intersections explain why SkillBrowser's pooled accuracy differs between comparator rows.

The pooled analysis treats model--instance pairs as observations. Repeated tasks across backbones may violate independence, so the nominal pooled $p$-values should not be interpreted as a hierarchical or cluster-adjusted analysis. Nor do they establish significance for every model--dataset setting. Per-setting sample counts and correctness outcomes are needed for 95\% Wilson accuracy intervals and stratified tests; the reported rounded accuracies alone do not determine those intervals.

\begin{table*}[t]
\centering
\footnotesize
\setlength{\tabcolsep}{4pt}
\renewcommand{\arraystretch}{1.12}
\begin{tabular}{@{}lrrrrrr@{}}
\toprule
Comparator & SB acc. & Ref. acc. & $N$ & $b$ & $c$ & $p$ \\
\midrule
Static \textsc{SkillGrep} & 78.59 & 74.64 & 11,736 & 1,110 & 647 & $1.46\times10^{-28}$ \\
LLM Selection & 75.93 & 72.82 & -- & 646 & 440 & $4.41\times10^{-10}$ \\
Gold Exposure & 78.59 & 80.52 & -- & 692 & 920 & $1.49\times10^{-8}$ \\
\bottomrule
\end{tabular}
\caption{Pooled valid-pair accuracy (\%) and two-sided exact McNemar tests. $b$/$c$ count exclusive SkillBrowser/comparator wins. Dashes denote paired sample sizes unavailable in the aggregate statistics; they must not be assumed equal across comparators.}
\label{tab:paired_statistics}
\end{table*}

\end{document}